\documentclass{naturep}

\usepackage{amssymb}
\usepackage{amsmath}
\usepackage{graphicx}

\usepackage{algorithmicx}
\usepackage{algorithm}
\usepackage[table]{xcolor}
\usepackage[noend]{algpseudocode}
\usepackage{bbm}
\usepackage{lineno}
\usepackage[margin=0.6in]{geometry}
\usepackage{ragged2e}
\usepackage{setspace}
\usepackage{longtable}
\usepackage{hyperref}
\usepackage{anyfontsize}
\usepackage{setspace}
\usepackage{float}
\usepackage{booktabs}
\usepackage{multirow}
\usepackage{xcolor} 
\usepackage{blindtext}
\usepackage{tabularx}
\usepackage{caption}
\usepackage{graphicx}
\usepackage{lineno}
\usepackage{appendix}
\usepackage{subcaption}

\usepackage[table]{xcolor}

\makeatletter
\let\saved@includegraphics\includegraphics
\AtBeginDocument{\let\includegraphics\saved@includegraphics}

\makeatother
\usepackage{listings}

\usepackage{enumitem, kantlipsum}

\definecolor{N}{RGB}{173,216,230}
\definecolor{ASCUS}{RGB}{152,251,152}
\definecolor{LSIL}{RGB}{255,182,193}
\definecolor{ASCH}{RGB}{221,160,221}
\definecolor{HSIL}{RGB}{255,236,139}
\definecolor{AGC}{RGB}{222,184,135}

\title{\begin{flushleft}{\begin{spacing}{1}Large-scale cervical precancerous screening via AI-assisted cytology whole slide image analysis\end{spacing}}\end{flushleft}}

\begin{document}

\maketitle

\begin{spacing}{1.8}
\vspace{-15mm}
\noindent Honglin Li$^{1,2\boldsymbol{\ddag}}$, Yuxuan Sun$^{1,2\boldsymbol{\ddag}}$, Chenglu Zhu$^{1}$, Yunlong Zhang$^{1,2}$, Shichuan Zhang$^{1,2}$, Zhongyi Shui$^{1,2}$, Pingyi Chen$^{1,2}$, Jingxiong Li$^{1,2}$, Sunyi Zheng$^{1}$, Can Cui$^{1}$, Lin Yang$^{1*}$
\end{spacing}
\vspace{-6mm}
\begin{spacing}{1.4}
\begin{affiliations}
 \item Research Center for Industries of the Future and School of Engineering, Westlake University
 \item Zhejiang University
 \\$\boldsymbol{\ddag}$ Contributed Equally
 \\\textbf{*Corresponding author}: Lin Yang (yanglin@westlake.edu.cn)
\end{affiliations}
\end{spacing}

\begin{spacing}{1.2}
\noindent \textbf{Cervical Cancer continues to be the leading gynecological malignancy, posing a persistent threat to women's health on a global scale. 
Early screening via cytology Whole Slide Image (WSI) diagnosis is critical to prevent this Cancer progression and improve survival rate, but pathologist's single test suffers inevitable false negative due to the immense number of cells that need to be reviewed within a WSI. 
Though computer-aided automated diagnostic models can serve as strong complement for pathologists, their effectiveness is hampered by the paucity of extensive and detailed annotations, coupled with the limited interpretability and robustness. These factors significantly hinder their practical applicability and reliability in clinical settings. 
To tackle these challenges, we develop an AI approach, which is a Scalable Technology for Robust and Interpretable Diagnosis built on Extensive data (STRIDE) of cervical cytology. STRIDE addresses the bottleneck of limited annotations by integrating patient-level labels with a small portion of cell-level labels through an end-to-end training strategy, facilitating scalable learning across extensive datasets. 
To further improve the robustness to real-world domain shifts of cytology slide-making and imaging, STRIDE employs color adversarial samples training that mimic staining and imaging variations.
Lastly, to achieve pathologist-level interpretability for the trustworthiness in clinical settings, STRIDE can generate explanatory textual descriptions that simulates pathologists' diagnostic processes by cell image feature and textual description alignment.
Conducting extensive experiments and evaluations in 183 medical centers with a dataset of 341,889 WSIs and 0.1 billion cells from cervical cytology patients--believed to be the largest dataset in this field to date--STRIDE has demonstrated a remarkable superiority over previous state-of-the-art techniques. This superiority is evident in various critical dimensions, including but not limited to, diagnostic accuracy, model robustness, and the pathologist-level interpretability.
}
\end{spacing}

\newpage

\begin{spacing}{1.35}
\noindent\textbf{\large{Introduction}}

Cervical Cancer remains the predominant gynecological malignancy and notably in 2020, it ranked as the second leading cause of Cancer-related fatalities among women aged 20–39\cite{cas2023}. Encouragingly, the emergence of Human-Papillomavirus (HPV) vaccines and advancements in cytology screening raises the prospect of Cervical Cancer evolving into one of the most preventable Cancers\cite{lancet_bonjor}. 
This is primarily because its onset is mainly attributed to HPV infections (according to WHO\cite{WHO}, 99\% Cervical Cancer cases are linked to high-risk HPV infections), thus the lesion development process of cervical epithelial cells can be observed via early Cytology screening or Colposcopy\cite{Burd2003,Okunade2019}, which is less traumatic compared to Histology tissue section or biopsy. Hence, Cervical Cytology (CC) early screening can be widely applied. 
The current thin-layer liquid-based cell smear technique has significantly enhanced the quality of CC sample, meanwhile the digital pathological slide scanning allows for its long-term preservation\cite{Piccione_Baker_2023}. These technological advancements have greatly improved the quality of the examination for traditional cervical smears.
However, achieving comprehensive Cervical Cancer screening in developing nations is especially challenged by their large populations and constrained medical resources, as noted in studies\cite{Bruni_Laia_2022,zhao2021analysis}. 
Additionally, the reliance on manual visual scanning by human pathologists may diminish the quality of the screening process when scaled up to serve vast populations. 
These limitations have been impeding the effectiveness and widespread adoption of CC screening programs.

In recent years, machine learning-based computer-assisted diagnosis, also known as AI diagnosis, has shown performance on par with that of well-trained doctors in various clinical applications, including medical CT, MRI, Microscopic images\cite{alaa2021machine,kaissis2021end,wang2021global,degrave2023auditing} and pathology image analysis\cite{binder2021morphological,wei2019pathologist,zhang2019pathologist,lu2021data,jain2020predicting}. These advancements are progressively underscoring the substantial clinical potential and value that AI brings to the field of diagnostics. 
For CC screening, the AI diagnosis could serve as a robust complement for pathologists as studied in\cite{cheng2021robust,zhu2021hybrid}, especially enhancing diagnosis efficiency and consistency when managing large-scale populations.
However, accurate prediction of healthcare outcome using AI is indeed challenging, primarily because AI models require extensive and high-quality training data and annotations to achieve optimal performance\cite{jia2023importance,day2023improving,steyaert2023multimodal}. 
There is a trend in computer vision tasks on natural images\cite{krizhevsky2012imagenet,he2016deep,xing2019towards,cai2021generalizing,zhang2022weakly} that increasingly rely on a large amount of labelled data to improve generalization\cite{kirillov2023segment}. However, in biomedical image analysis\cite{jia2023importance,liang2023deep,lutnick2019integrated}, extensive and detailed annotations are infeasible due to the complexity of medical data and scarcity of professional doctors. This challenge is particularly exacerbated in the context of CC Whole Slide Image (WSI), which typically contains an extraordinarily large number of cells—on average, approximately 80,000 cells per WSI. As a result, it is tedious for pathologists to annotate all these cells, and an more practical approach is to obtain only a small portion of cell-level labels ( or sparsely annotated subset of the WSIs) and obtain weak labels at the WSI-level (patient-level).

When applying the simple supervised-learning method to CC WSI analysis tasks based on the weak annotations, the performance often falls short of real-world needs. Previous efforts to address weakly supervised learning for CC WSI typically involve multi-stage training with given annotations for each stage. 
For instance, Cheng et al.\cite{cheng2021robust} propose a 3-stages method: initially detect all lesion-positive cell nuclei via heatmap key-points\cite{xing2019towards, cai2021generalizing}, then retrain classifier with ResNet\cite{krizhevsky2012imagenet,he2016deep} to make fine-grained classification on detected cells, finally pick top-10 cells and input their feature representation of ResNet into an RNN model\cite{campanella2019clinical} to make WSI diagnosis. 
Another method proposed by Zhu et al.\cite{zhu2021hybrid} follows a similar paradigm but uses YOLO-v3\cite{redmon2018yolov3} as the cell detector and Xception\cite{chollet2017xception} for cell classification, then employs Xgboost\cite{chen2016xgboost} to make WSI-level diagnosis with selected features.
However, these methods are constrained by the use of labelled data only, thus lacks scalability when applied to larger datasets without detailed annotations, preventing the learning of generalized representations for real-world clinical applications. Therefore, they often show limited performance and robustness.
A similar problem also arises in Histology WSI analysis\cite{wu2021radiological,prosperi2020causal,jain2020predicting,faust2019intelligent,zhang2019pathologist,carrillo2024generation} where typically only slide-level labels are accessible. Various studies\cite{campanella2019clinical, lu2021data, jaume2022integrating} treat WSI as a sequence of patches (or a bag of patch instances), employing  Multiple Instance Learning (MIL)\cite{campanella2019clinical} with Attention-based pooling\cite{ilse2018attention}.
However, owing to the extremely high resolution of WSIs, these studies also employ separated-stages training approaches: a frozen visual backbone pre-trained on the ImageNet\cite{russakovsky2015imagenet} is used to extract fixed features for each patch instance, followed by the application of MIL on all instances' features to derive the final prediction.
Given the vast quantity of unlabelled patch images, efforts to learn better pre-trained features for the pathology domain, without extensive annotations, have been conducted by\cite{li2021dual, chen2022scaling, sish, xu2024whole, chen2024towards} through the implementation of visual Self-supervised Learning (Self-SL)\cite{MOCO, chen2020simple, caron2021emerging, MAE, krishnan2022self}. Subsequently, they conduct linear probing\cite{chen2020simple, MOCO} on WSI level (freeze all backbone then train the classifier head).
However, employing linear probing following Self-SL is sub-optimal compared to full fine-tuning on downstream tasks in both vision\cite{MOCO, chen2020simple, caron2021emerging, MAE,wang2020augmenting,azizi2023robust} and language\cite{devlin2018bert,floridi2020gpt} tasks.
Another approach for utilizing vast quantities of unlabelled data is Semi-supervised Learning (Semi-SL)\cite{tarvainen2017mean, berthelot2019mixmatch, sohn2020fixmatch}, which generates pseudo labels or learns consistency of different augmentation views. Nonetheless, in real-world applications, particularly in analyzing CC WSIs, the imbalance distribution of categories leads pseudo labels being biased towards predominant categories\cite{wei2021crest, kim2020distribution} during Semi-SL, thus strongly hindering the classification performance.

We argue that the existing separated-stages training approaches are sub-optimal for CC WSI analysis due to a fundamental misalignment between the objectives of cell-level feature pre-training and the task of the WSI classification. This is inevitable whether by supervised learning on small portion cell annotations (under-fitting), ImageNet transfer learning (domain-gap), Self-SL (pre-training and fine-tuning objective gap), or Semi-SL (increased categories distribution bias). What's more, the cell-level representation can not be scaling up on more WSI weakly-supervised data, limiting the performance and robustness.
An intuitive approach to smooth this problem is further fine-tuning the cell-level representation by WSI level label information, aka, end-to-end learning. 
However, due to the large volume of cells within a WSI, previous works\cite{campanella2019clinical, lu2021data, zhu2021hybrid, cheng2021robust} fail to accomplish this goal to compromise training efficiency.
Motivated by the Max-pooling nature of MIL and the low-rank property of WSI instances (see Method for detail), we propose to learn the cell backbone and WSI head in an end-to-end representation learning paradigm by distilling top-K cell instances of a WSI as a bag for efficiency. 

We initially adopt the traditional approach used in previous studies\cite{cheng2021robust,zhu2021hybrid}, which extracts the top-K positive-like cells by training a simple but efficient YOLO cell detector\cite{redmon2018yolov3,bochkovskiy2020yolov4,jocher2020yolov5}.
To enhance cell representation learning, we apply Semi-SL with a set of annotated cells and the detected unlabelled top-K cells from the WSI. Differing from previous methods that aimed to minimize category bias by re-sampling\cite{wei2021crest, kim2020distribution}, we address this inherent challenge of Semi-SL by leveraging WSI-level label information to refine the pseudo labels of top-K cells. 
By employing this learning scheme of our STRIDE-AI, the advantages are in various folds:
1) We can effectively leverage the label information from WSI, ensuring that the learned cell model representations are better aligned with the objectives of WSI diagnosis. 
2) It help us scaling up the learning on extensive data with only WSI weak label for more generalized feature, enhancing the model performance. 
3) The cell-level image augmentations is plug-and-play given end-to-end training, especially color adversarial samples that we proposed to simulate staining differences or domain shifts\cite{madan2022and} can further bolstering the robustness.
4) In addition, for the pathologist-level interpretability\cite{zhou2022generalized,zhang2019pathologist}, we annotate each cell diagnosis with explanatory textual descriptions to illustrate the rationale behind the diagnoses, mimicking pathologists' diagnostic processes, where the more generalized visual feature learned by prior process can help learning better alignment between image and textual descriptions. 

\begin{figure*}  
\centering
\includegraphics[width=1\textwidth]{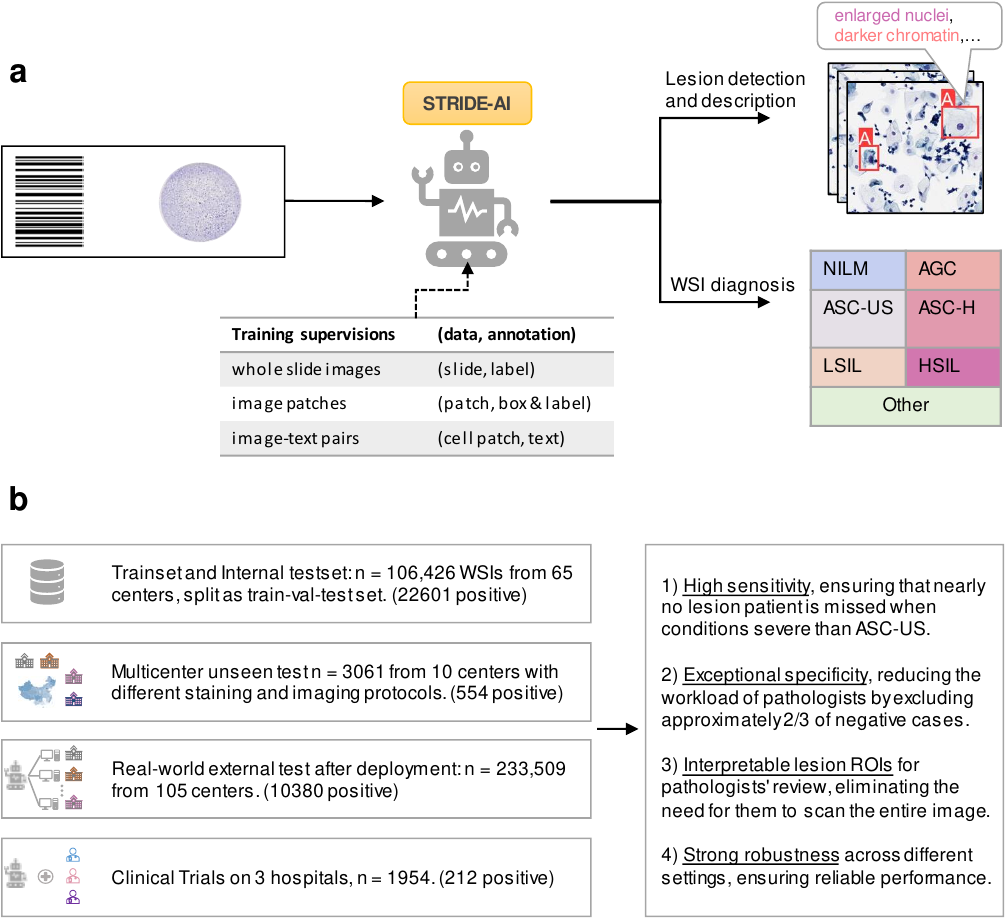} 
\caption{\textbf{Overview of the model development and evaluation. a.} Model development. STRIDE takes digital slide of Cervical cytology as input and outputs the WSI-level diagnosis probability and the detected lesions cells with corresponding textual description; STRIDE was trained with confirmed patient-level WSI labels and cell-level box, category and textual labels. \textbf{b.} Data collection and evaluation for model. We evaluate the performance of STRIDE on the internal test data, multicenters unseen test centers with staining variation, real-world external test with multicenters after model deployment and clinical trial assisting pathologists' diagnosis.
}  
\label{fig_datasets1}
\end{figure*}

To develop and evaluate our method, extensive experiments are performed on a large-scale of data including 183 data centers with 341,889 WSIs (as shown in \textbf{Fig. \ref{fig_datasets1}b} and \textbf{Extended Data Fig. 1b}), demonstrating that STRIDE notably surpasses existing state-of-the-art techniques in terms of accuracy and robustness in both internal and real-world external testset. 
The development dataset for training is meticulously annotated and curated by pathologists through our custom web-based annotation programs, following rigorously designed procedures. For a detailed overview of the data preparation process and a concise summary of dataset information, please refer to \textbf{Extended Data Fig. 1}.
The significant challenge in the real-world application of CC WSI diagnosis is the variation in staining and imaging protocols across different hospitals or data centers. To assess the robustness of STRIDE to such domain-shift, we validate it on 10 centers with unseen domains of 3061 WSIs data. 
This validation set is crucial for real-world AI deployment, an aspect where previous works\cite{cheng2021robust,zhu2021hybrid} ignored but struggled to generalize.
We further make real-world external test after our model deployment with 233,103 WSIs from 105 centers and clinical trial on 1954 participated patients form 3 top-tier hospitals, showcasing its potential for widespread clinical application.
The comprehensive evaluations support our STRIDE-AI with ability including high sensitivity, exceptional speicificity, interpretable lesion cells region of interest (ROI) and strong robustness (\textbf{Fig. \ref{fig_datasets1}b}), thus showing the promising potential for assisting human pathologists' diagnosis with efficiency and effectiveness.

\clearpage
\noindent\textbf{\large{Results}}

\noindent\textbf{Overview of STRIDE}\\
The objective of STRIDE is to automatically provide accurate, interpretable, and robust CC diagnosis. This approach not only consistently assists pathologists in excluding a substantial proportion (about 2/3) of slides from lesion-negative patients, but also reduces missed diagnosis rates for positive lesions. In contrast to previous methods\cite{zhu2021hybrid,cheng2021robust} that focus on fully supervised training with limited annotations and data, STRIDE introduces an annotation-efficient training workflow. This innovative approach enables the acquisition of more comprehensive knowledge by incorporating diverse data from various hospital domains, enhancing scalability and generalization to real-world big data distributions. 
Furthermore, STRIDE possesses the unique capability to generate explanatory textual descriptions, mirroring the diagnostic process followed by pathologists and aligning with TBS\cite{nayar2015bethesda} definitions. Consequently, it can serve as a more reliable tool for clinical routine use.

\begin{figure*}
\centering
\includegraphics[width=\textwidth]{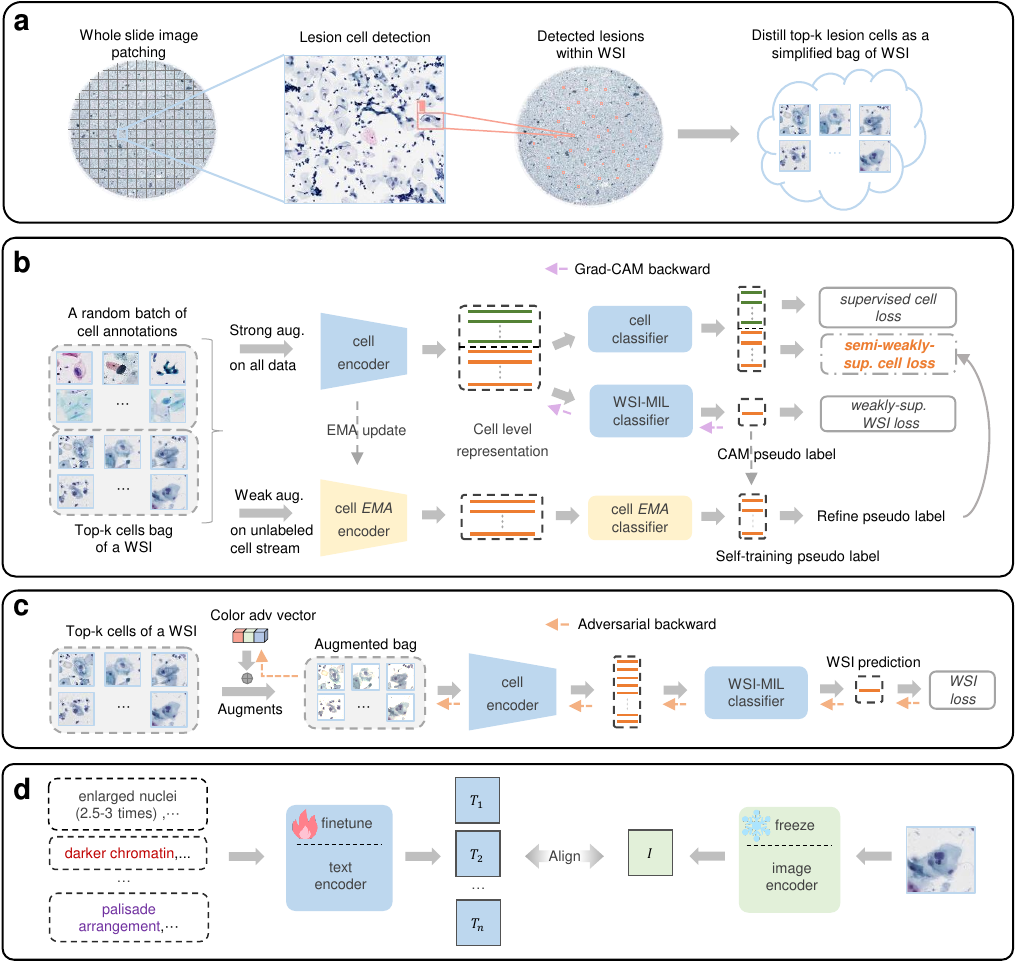} \caption{\textbf{Overview of STRIDE. a.} Top-K cells distillation of WSI. The foreground regions of WSI are patched and  then perform lesion-positive cell object detection to select top-K representative cell patches. 
\textbf{b.} The model structure and SWIFT training scheme. We first train cell classifier with all annotated cells, then fine-tune the cell representation backbone and the WSI classification head in an end-to-end way by combining the Semi-weakly supervised integrated fine-tuning, where the labelled cells comprise the supervised stream and the unlabelled top-K cells of a WSI are the weak-supervised stream. The supervised stream is trained via strong augmentation in a traditional way, while in the meantime,  the weak-supervised stream is trained via two components: 1) the WSI level end-to-end MIL training; 2) the cell level Semi-weakly supervised training. 
\textbf{c.} The ColorAdv steps: color adversarial example generation and generalized model learning, which are alternatively optimized. The first step involves identifying color adversarial examples by maximizing the loss, while the second step involves updating the model using a mixture of generated color adversarial examples and original examples. 
\textbf{d.} The cell image and text description feature alignment. The visual cell classification model is frozen since it captures morphological features by former training steps. A language model encoding the interpretable text into feature vectors is fine-tuned to align the text features with the visual features.}
\label{fig_framework}
\end{figure*}

Algorithmically, the training process of STRIDE follows an mixed-training procedure, comprising pre-training, partial-label tuning, and weakly-supervised tuning. The architecture scheme is illustrated in \textbf{Fig. \ref{fig_framework}}. Distinct from conventional approaches\cite{zhu2021hybrid,cheng2021robust}, STRIDE encompasses three key advantages:
\textbf{1) Annotation efficiency for improved accuracy (Fig. \ref{fig_framework}a+b)}: This feature enables the model's ability to address and comprehend the complexities inherent in large-scale datasets. 
To boost training efficiency, a straightforward lesion-positive cell object detector is utilized to identify the most suspicious cells from a WSI, selecting the top-K (e.g., K=1024) cell instances for each WSI, gathering a total of 0.1 billion (1024*100k) cells for self-supervised pre-training and further weakly-supervised fine-tuning. The selection process for the top-K cells aligns intuitively with the Max-pooling and Multiple Instance Learning (MIL) framework used in WSI diagnosis\cite{campanella2019clinical} (see \textbf{Method} for details). We initially pre-train the cell backbone with all cell annotations derived from the top-K cells of each WSI.
Subsequently, the cell representation backbone and the WSI classification head are fine-tuned end-to-end within the MIL framework and integrated with \textbf{S}emi-\textbf{W}eakly supervised \textbf{I}ntegrated \textbf{F}ine-\textbf{T}uning (\textbf{SWIFT}), enabling scalability in learning from more WSIs data. 
\textbf{2) Robustness to staining changes and imaging variations by designed augmentations (Fig. \ref{fig_framework}c)}: This aspect facilitates the integration of various augmentations during the training of WSI models. Additionally, color adversarial training dynamically simulates real-world staining changes, enhancing the model's robustness against unseen hospital data. It's essential to emphasize that these image-level augmentations can only be performed under the end-to-end tuning framework of SWIFT.
\textbf{3) Enhanced Interpretability (Fig. \ref{fig_framework}d)}: The STRIDE system is capable of generating explanatory text descriptions by leveraging the morphological features observed in cells. This capability stems from its training phase process, where the visual attributes of cells are aligned with their respective textual annotations, which closely mirrors the diagnostic process of pathologists as defined by the TBS\cite{nayar2015bethesda} system, thereby making its diagnostic outputs more intuitive and enhancing the interpretability of the STRIDE system.

\noindent\textbf{Cell detection and classification refinement}\\
For cell detection, the results are listed in \textbf{Extended Data Fig. 4a}. The mAP@50-95 (COCO\cite{COCO} metric) of the YOLO-v5m model reaches 48.4\% and we find no strong improvement in 20x magnification. The recall (at threshold=0.001) reaches 0.989, which implies that top-K recall of a positive WSI can hardly be zero in WSI of multi positive cells, e.g. if there are 2 positive cells within a WSI, the miss-rate of all positive cells will be $(1-0.989)^2 = 0.000121 = 0.0121\%$ . The cells detector can also alleviate small label noise (when pathologists miss some positive cells in a patch), demonstrating its strong potential for down-stream cell classification and WSI analysis task use.

After detecting lesion suspected cells, we aim to learn a better cell-level feature for the downstream cell and WSI-MIL multi-classification tasks. 
The advantage of this refinement stage is that we can form a cascade structure since the relatively low-accurate cell detector with lower input image magnification (at 10x) and smaller model size are good enough to filter out the most irrelevant negative cells. Then, with larger input magnification (at 20x) and model size at this stage, the cell classifier can concentrate more on the detected top-K hard samples of WSIs thus boosting the accuracy.
\textbf{Extended Data Fig. 1d} provides a visual representation of the different classes, which shows some inter-class similarity especially in adjacent classes. Considering the subtle distinction at the cell level among SCC, HSIL and ASC-H, we combine them into a single category to mitigate model confusion. 
To address the variability of the multi-center dataset and improve overall performance, we employ a ConvNext-tiny model\cite{liu2022convnet} for fine-grained classification and learning on all available annotated multi-centers of cell data. The evaluation of the model verifies its effectiveness, attaining a F1-score of 78.34\% (\textbf{Extended Data Fig. 5}).

\begin{figure*}
    \centering
    \includegraphics[width=\textwidth]{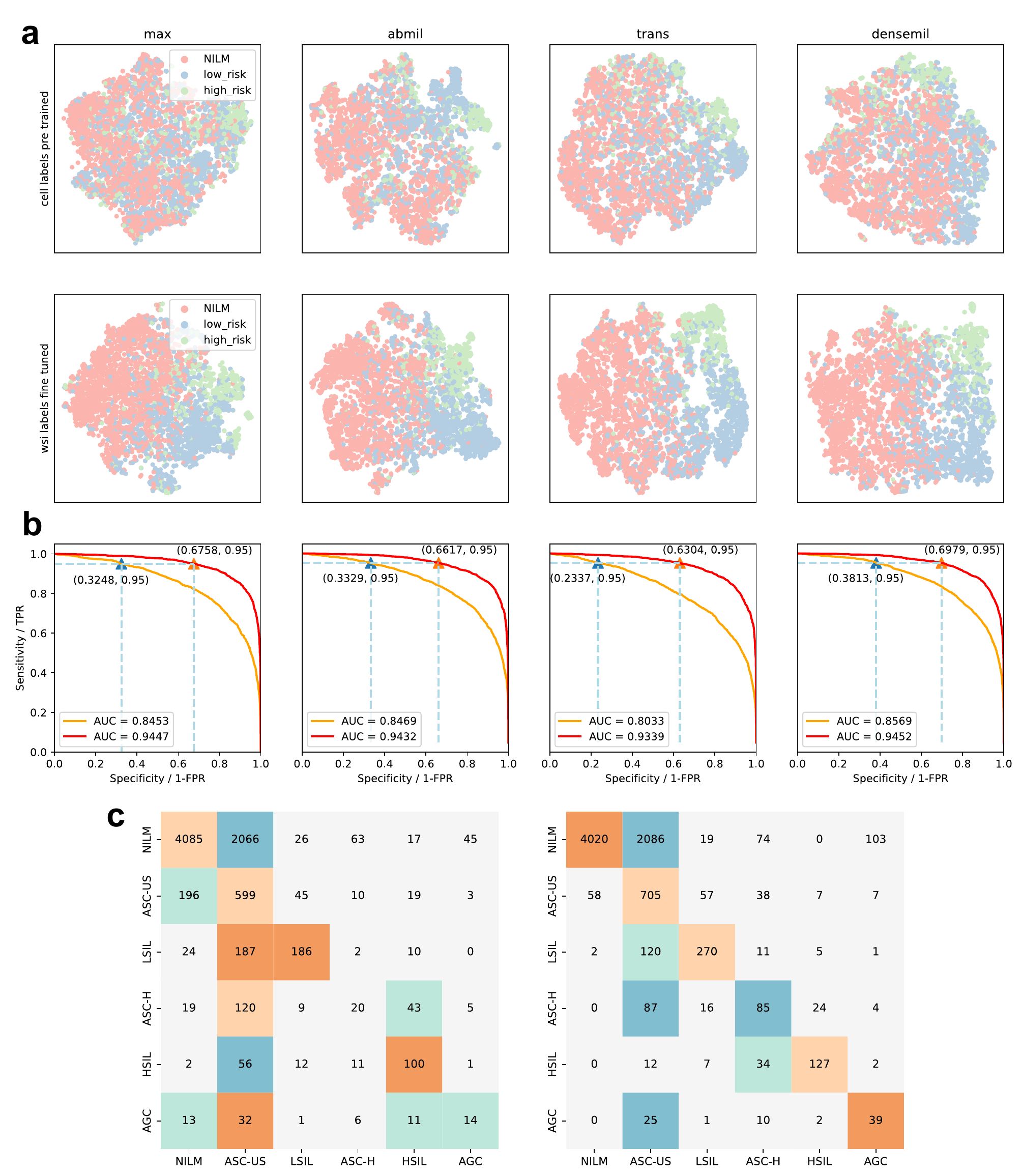}  
    \caption{\textbf{WSI diagnosis results with different WSI-head architectures and cell level features.} \textbf{a.} WSI feature T-SNE decomposition and AUC details on different WSI-head with different cell level embeddings (cell-level labels supervised pre-trained and WSI-level labels fine-tuned). \textbf{b.} The corresponding binary classification ROC curve. \textbf{c.} WSI confusion matrix comparison between cell level annotation pre-trained (left) and WSI-level fine-tuned (right) embedding, given similar specificity (0.65).}
    \label{fig_wsi_res_arch}
\end{figure*}

\begin{figure*}
    \vspace{-15mm}
    \centering
         \includegraphics[width=0.9\textwidth]{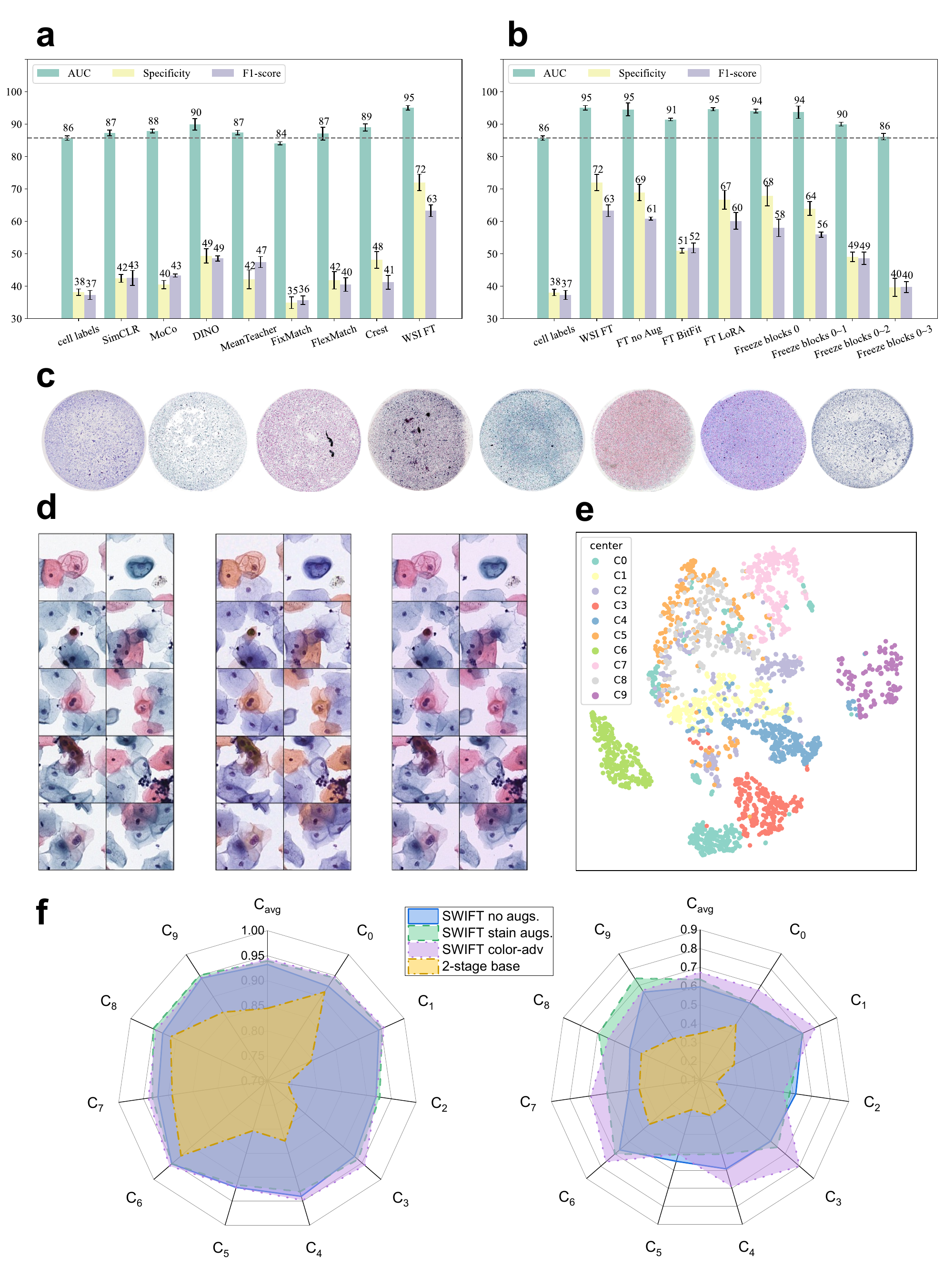}
\caption{ \textbf{WSI diagnosis performance and robustness on internal test data.} \textbf{a.} Semi-/Self- supervised learning features on given cell annotations and unlabelled top-K cells of WSIs. \textbf{b.} WSI top-K fine-tuning (WSI-FT, or SWIFT) features with/without cell-level augmentations and different fine-tuning tricks.  \textbf{c.} Staining color changes among different data centers. \textbf{d.} The first column are original images and the remain two columns are color adversarial samples in RGB and HSV space. \textbf{e.} The corresponding staining distributions. \textbf{f.} The performance on domain-shift, the left and the right one are AUC and specificity respectively.}
    \label{fig_wsi_res_features}
\end{figure*}

However, since there are only less than 3.5\% positive cells are annotated (as mentioned in data challenge in \textbf{Data} section, forming an incomplete sampled subset of all cells from all the 65 centers of WSIs), we find that the learned cell feature representation can only get a relatively low WSI-level performance (\textbf{Fig. \ref{fig_wsi_res_arch}}). 
The sparse cell annotation problem is realistic but quite challenging, which motivates us to incorporate more unlabelled data and WSI-level label information. 
We first incorporate unlabelled cell data by Semi-SL and Self-SL + fine-tuning, but they still fail in WSI classification \textbf{Fig. \ref{fig_wsi_res_features}a}.
We then perform the SWIFT by adding unlabelled top-K cells with WSI level annotation information, enabling the model to capture more intricate details and subtle distinctions among the large-scale cell distribution. 
The evaluation of the model demonstrates better effectiveness, achieving an impressive F1-score of 80.89\% (\textbf{Extended Data Fig. 5}). 
With the refined classification model at our disposal, we can effectively detect and classify cells, while extracting valuable classification features and probability information for further processing and analysis like WSI diagnosis, resulting in better WSI-level results (\textbf{Fig. \ref{fig_wsi_res_features}a+b}).

\clearpage
\noindent\textbf{Whole Slide Image diagnosis}\label{wsi_exp_res}\\
The ultimate goal of this study is to achieve a comprehensive diagnosis of WSIs in accordance with the TBS\cite{nayar2015bethesda} diagnostic criteria. 
Common squamous intraepithelial lesions, including ASC-US, LSIL, ASC-H, HSIL+SCC, and gland lesion (AGC), as same to cell level categories, must be identified and classified. These specific lesion types are crucial in determining the diagnosis and guiding treatment protocols. 
In order to accomplish this challenging task, we leverage Attention-based WSI-MIL\cite{ilse2018attention} model, incorporated with densely-connected MLPs to effectively differentiate positive and negative lesions. 
The performance of our MIL model is illustrated in two main aspects: 1) WSI-MIL architecture (\textbf{Fig. \ref{fig_wsi_res_arch}a+b}) and 2) cell backbone representation learning for WSI diagnosis (\textbf{Fig. \ref{fig_wsi_res_arch}, Fig. \ref{fig_wsi_res_features}a+b}), both showing superiority over various competing baselines. 

For WSI-MIL architectures, we compare their performance on both cell pre-trained features and SWIFT features, which shows apparently improvement of our architecture. However, we find that none of the WSI-MIL models can get acceptable performance (with a AUC $<$ 90, and specificity@sense95 $<$ 50\%) with only cell supervised learning. 
Thus we argue that, with supervised training on limited cell samples, the learned feature representation is not effective for WSI diagnosis. 
For cell backbone features representation learning, despite our SWIFT feature, we also compared the features learned by visual Self-SL, but found no significant superiority since their training objective is far from downstream tasks such as cell and WSI recognition. 
The Semi-SL is introduced for its benefits of both supervised and unsupervised training, but it is fragile to categories' imbalance and shows limited improvements in our setting. 
Notably, by further enhancing the backbone with WSI-level annotation information via SWIFT, the learned representation demonstrates a remarkable specificity of 72.02\% at a sensitivity of 95\%, along with an impressive AUC score of 95.01. 
It is noteworthy that almost all the misclassified 5\% positive WSIs are labelled as ASC-US (right column of \textbf{Fig. \ref{fig_wsi_res_arch}c}), which is acceptable considering the inherent challenges of its feature space is very close to negatives (\textbf{Extended Data Fig. 1d}). 
In conclusion, our system exhibits high sensitivity and exceptional specificity, in line with the primary goal of CC computer-aided diagnosis. These false-positive WSIs (less than 1/3 in total) are subjected to further review and scrutiny by experienced pathologists to ensure accurate diagnosis and subsequent treatment decisions. The comprehensive analysis and evaluation indicate that our WSI analysis system holds great promise, as it facilitates the pre-screening of slides without apparent abnormalities, thereby reducing the workload of pathologists and improving overall diagnostic efficiency.

\noindent\textbf{Robustness to domain-shift} \label{color_adv_exp_res}\\
We also focus on addressing the challenge of domain-shift robustness in real-world pathological applications, since it is common for different hospitals or clinical centers to employ varying parameters for slide staining and making or utilizing different microscopic imaging devices. To evaluate the robustness of our model in this scenarios, we intentionally reserve data from 10 unseen hospitals.
Previous methods, as exemplified by\cite{campanella2019clinical}, use fixed patch/cell features (learning from limited cell annotation)s to train the WSI head to accelerate training, which hinders the generalization of the model. In contrast, our proposed method addresses these limitations by incorporating versatile training-time augmentations to the introduced fine-tuning scheme. This combination allows our model to achieve better generalization performance across various real-world domain shifts. 

Common image augmentations can provide about 3\% improvement of specificity and 0.5\% improvement of AUC on the internal testing set (\textbf{Fig. \ref{fig_wsi_res_features}b}). 
For the reserved unseen 10 centers of data, the large margin improvement ((\textbf{Fig. \ref{fig_wsi_res_features}f})) are comprised by two factors: 1) the end-to-end training can boost specificity improving 25\% since the model-backbone are learned on a more wide-spread data. 2) Our further proposed color adversarial training (\textbf{Fig. \ref{fig_framework}c} and \textbf{Method}) with stain augmentations, achieve averagely 67\% specificity on unseen data, compared to previous 2-stage methods baseline with only 35\% specificity. 
By explicitly considering domain-shift robustness and emphasizing the importance of training-time augmentations, our study contributes to the development of models that can effectively adapt to variations in data acquisition and imaging protocols. The evaluation of our model on unseen centers provides valuable insights into its performance in practical settings, highlighting its potential for real-world clinical application.


\begin{figure*}
    \centering
    \vspace{-10mm}
         \includegraphics[width=0.9\textwidth]{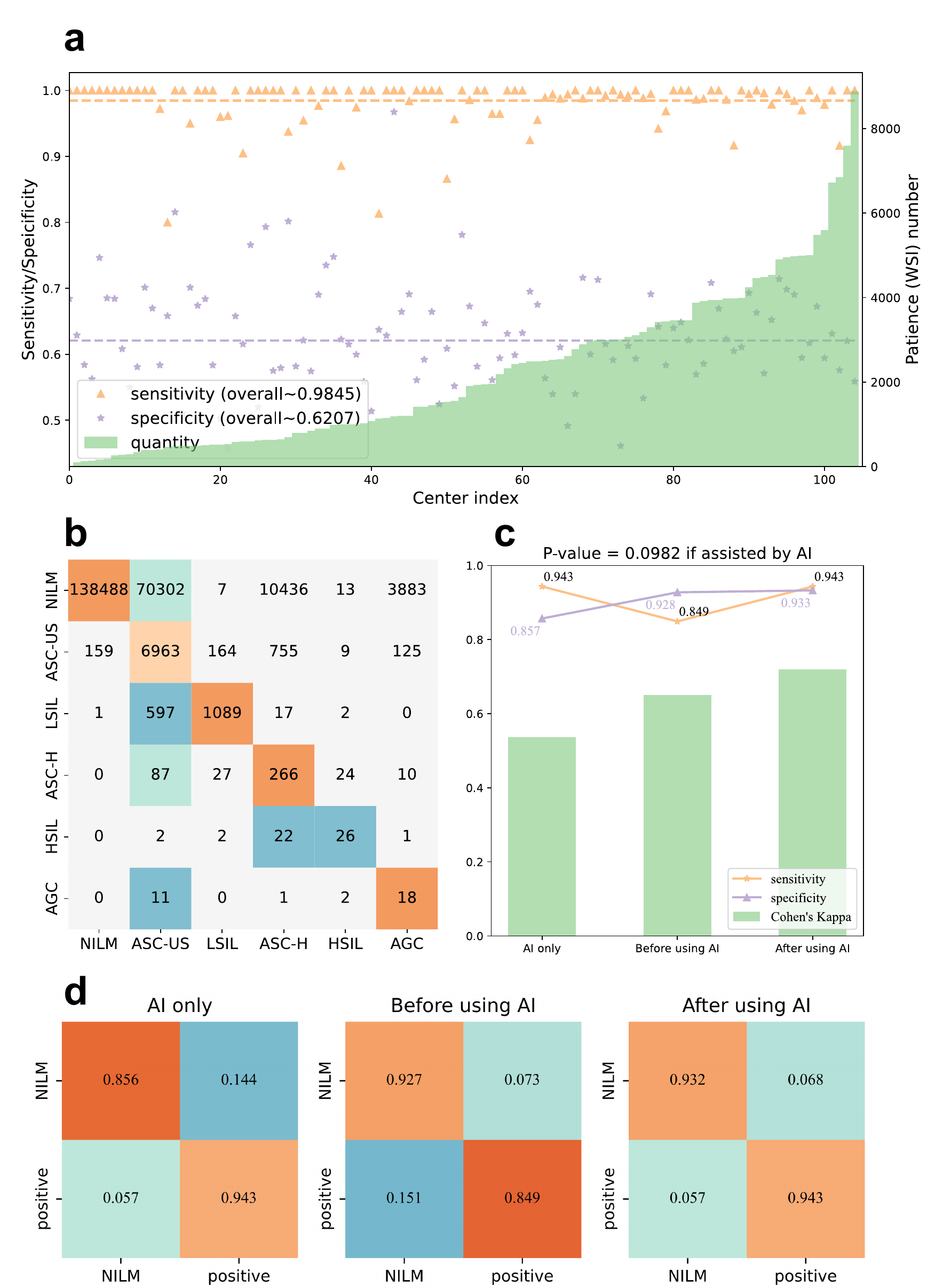}
         
\caption{ \textbf{WSI diagnosis performance on external test and clinical trial data.} \textbf{a.} Our model reaches high sensitivity and specificity in 105 unseen data centers with a total n = 233,509. \textbf{b.} The corresponding confusion matrix of (a). \textbf{c.} The clinical trial significant improvement: assisted with the AI, pathologists can improve both the sensitivity and specificity, thus a higher consistency (Cohen's Kappa = 0.7197) with gold standard (annotated by a more expert pathologists panel). \textbf{d.} The binary confusion matrix of our AI model and it's assistance result to pathologist. }
    \label{fig_5_external}
\end{figure*}

\noindent\textbf{Real-world external test and clinical trial} \label{clinical_trials}\\
To validate our model in real-world and clinical setting, we deploy our model to different centers and hospitals. The validated data is illustrated as \textbf{Fig. \ref{fig_datasets1}b} and \textbf{Extended Data Fig. 1}. We claim that our model can help pathologist's diagnosis for both developing township-level (resource-constrained region) and developed cities' hospitals.

For the real-world external data tested on 105 different centers (mainly from township level hospitals), the patients' quantity accounts ranging from 100 to 8500, with a total number of 233,509 and positive number of 10380. We list the data distribution, the sensitivity and specificity of each center in \textbf{Fig. \ref{fig_5_external}a}. The overall sensitivity = 0.9845 and specificity = 0.6207 is quite promising given that there is no further parameter tuning after our model being deployed, and this is close to the testing result during our model development. More detailed testing confusion matrix are depicted in \textbf{Fig. \ref{fig_5_external}b}, where we can observe that only 1 LSIL patient is missed from those lesion condition severe than ASC-US.

For the clinical trials on 3 hospitals (top-tier hospitals from highly-developed cities), we mainly evaluate how much our AI model can assisting human pathologist for better performance.
The trial's workflow is listed in \textbf{Extended Data Fig. 2}, mainly including participated patient and slide imaging control, controlled trial on if assisted by our AI model, third-party expert for gold standard labelling and data statistical analysis. 
The diagnosis result of human pathologist expert showing significant improvement after using our AI tool (\textbf{Fig. \ref{fig_5_external}c+d}), the sensitivity and specificity are both improved by a large margin, also with a higher consistency (approximated Cohen's Kappa = 0.72) to gold standard. 
For each hospital's clinical trial details, we show the confusion matrix in \textbf{Extended Data Fig. 3}. One of the trials' records are partially listed in \textbf{Extended Data Table. 1} with main findings that: 1) There is no lesion patient is missed by both AI and human reviewers when lesion conditions severe than ASC-US. 2) After arbitration, some lesion-positive slide (with low-grade or atypical lesion) suspected by human reviewers are determined as NILM (negative) diagnosis, this inconsistency between pathologists with varying professional experience and skill highlighting the need of AI-assistance with consistency and being stable.

\noindent\textbf{Explainable cell classification} \\
\begin{figure*}
\centering
\includegraphics[width=\textwidth]{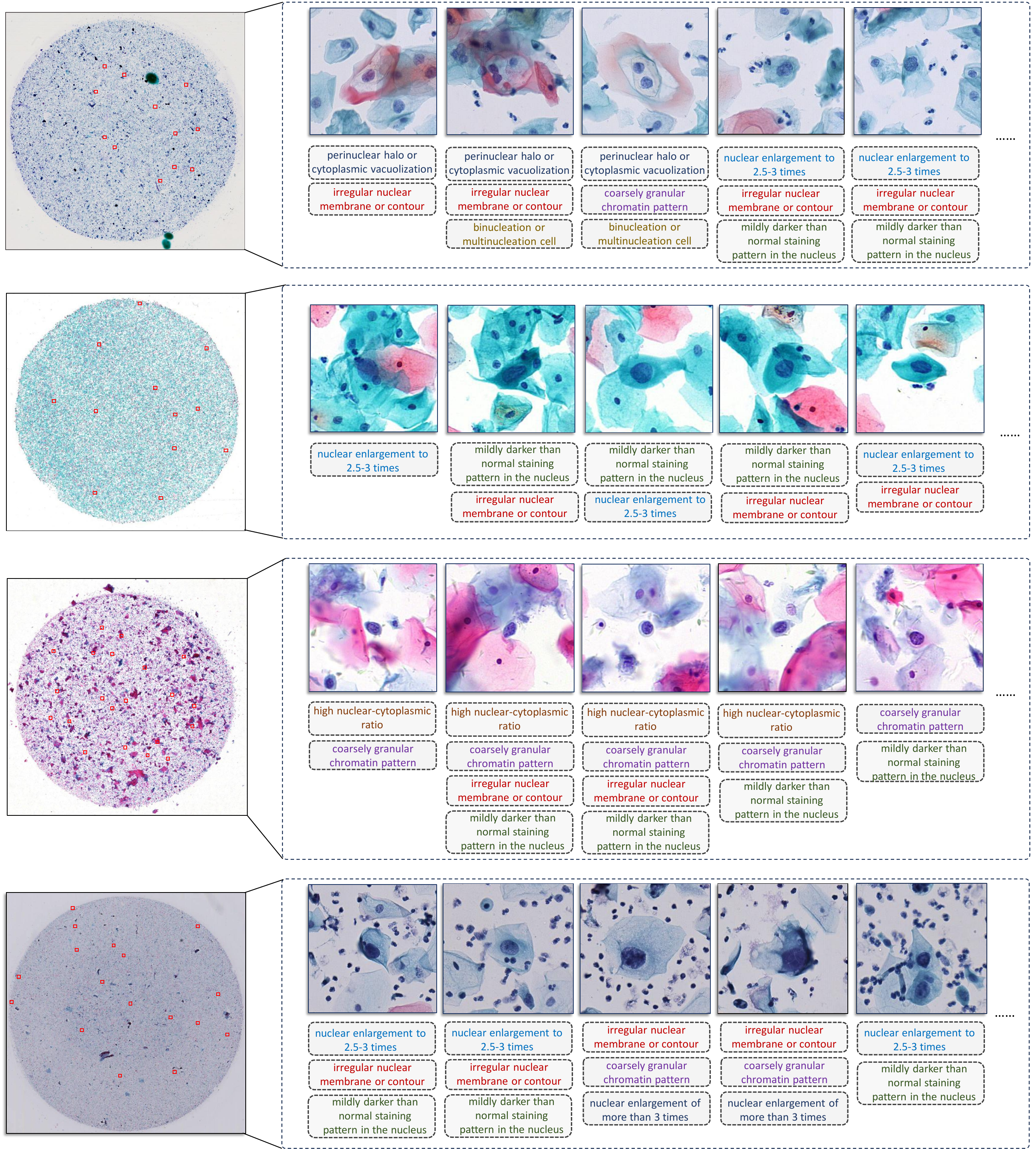}  
\caption{Examples of interpretable classification of cells in WSI.}  
\label{fig_cells_method}
\end{figure*}

In CC diagnosis, experts often rely on certain morphological changes in cells when classifying them, such as perinuclear halos or cytoplasmic vacuolization, which are typical characteristics of Low-grade Squamous Intraepithelial Lesion (LSIL) cells. Previous AI-driven approaches to cell classification have primarily focused on the direct identification of cell categories, while we remain unaware of the specific reasons behind the model's classifications. However, by concurrently providing the diagnostic basis along with the cell type, the system's interpretability can be significantly enhanced. This enhancement not only contributes to a better understanding but also positions the system as a potential pedagogical tool for training frontline pathologists.

To endow our system with this capability, we have engaged pathologists to comprehensively compile the potential criteria description used for determining cell categories. In order to streamline the annotation process, diagnostic descriptions are presented through a text-based multiple-choice format. These descriptions effectively capture the morphological characteristics of the cell image, as illustrated in \textbf{Extended Data Fig. 1c+d}, and detailed statistical information is provided in \textbf{Extended Data Table. 2}. During the process of annotating cell types, pathologists are tasked with selecting the descriptions that correspond to the diagnostic basis. Note that due to the limited number of Atypical Glandular Cells (AGC) present, there are fewer samples available for each description term specific to AGC. Consequently, for AGC cells, we consolidate all descriptors into ``atypical glandular cells".

The visual cell classification model (ConvNext-tiny) used in former sections, is already equipped with the logic to classify cells based on morphological features. However, this logic is only implicitly embedded within the visual feature vector. Therefore, we introduced a language model to encode the interpretable criteria into vectors, aligning these with the visual features for training. During the inference stage, given the visual feature of a cell, we simply select the texts with a similarity larger than a predefined threshold $\tau$ as its diagnostic explanation. The top-K ROI on WSI  with interpretable text (\textbf{Fig. \ref{fig_cells_method}} and \textbf{Extended Data Fig. 6}) will benefit much for the pathologists' further reviewing, and also makes our AI model trustworthy.

\clearpage

\noindent\textbf{\large{Discussion}} \\
Cervical Cancer poses a threat to women's health worldwide. However, developing nations face challenges in implementing comprehensive screening due to large populations and limited medical resources. Manual visual scanning by pathologists also reduces the efficiency and quality of large-scale screening. AI-based diagnosis systems have shown promise in various medical imaging applications, including Cervical cytology screening, by enhancing diagnosis efficiency and consistency. AI models, however, require extensive, high-quality training data and annotations. Recent advancements\cite{devlin2018bert,floridi2020gpt,kirillov2023segment,rombach2022high,radsch2023labelling} in AI research underscore the pivotal role of training models on larger datasets for achieving broader generalization\cite{gero2023incentive}.
Notable examples in natural language processing, such as BERT\cite{devlin2018bert} and GPT\cite{floridi2020gpt}, have demonstrated exceptional performance by leveraging Self-Supervised Learning (Self-SL) techniques, including masking-recovering or auto-regressive methods, on extensive text corpora. However, methods for vision tasks, like SAM\cite{kirillov2023segment}, and Diffusion models\cite{rombach2022high}, often rely on a substantial amount of annotated data. In the context of medical image AI models, the stochastic annotation process\cite{radsch2023labelling} is commonly employed, limiting the scalability of the model on large-scale datasets. Biomedical image analysis faces unique challenges due to complex medical data and limited professional annotators. In Cervical cytology whole slide images (WSI), the large number of cells per slide makes comprehensive annotation impractical. Thus, using weak annotations at the WSI level combined with limited cell-level labels is a more feasible approach.

Simple supervised learning with weak annotations often underperforms in real-world applications. Previous efforts involve multi-stage training methods, but these are limited by the need for detailed annotations, reducing their scalability and robustness (\textbf{Fig. \ref{fig_wsi_res_features}a+f}). Similar issues exist in histology WSI analysis, where slide-level labels are common, and studies use methods like multiple instance learning (MIL) with attention-based pooling. However, these methods also rely on pre-trained visual features, which may not align well with WSI classification tasks.
A practical solution is to fine-tune cell-level representations using WSI-level labels, achieving end-to-end learning. This approach is more scalable and can handle large datasets with weak labels. Our proposed method, STRIDE-AI, utilizes Semi-supervised Learning with class activation map of WSI-level labels to distill pseudo-labels of cells, resulting in accurate cell-level predictions. This approach improves the robustness and efficiency of Cervical cytology screening, making it a promising solution for resource-constrained regions (\textbf{Fig. \ref{fig_wsi_res_features}} and \textbf{Fig. \ref{fig_5_external}}). To better assist the decision-making of pathologists, textual description is given for AI diagnosis by multi-modality learning (\textbf{Fig. \ref{fig_cells_method})}.

Our STRIDE-AI system offers multiple benefits: 1) Aligns cell model representations with WSI diagnosis objectives. 2) Scales learning to extensive datasets with WSI weak labels, improving model performance. 3) Incorporates plug-and-play cell-level image augmentations, enhancing robustness. 4) Provides interpretability with explanatory textual descriptions for each cell diagnosis.
Extensive experiments and evaluation on real-world data and clinical settings have demonstrated STRIDE-AI with high sensitivity, specificity, and strong robustness, highlighting its potential for widespread clinical application and assisting human pathologists with efficient and effective diagnosis.
In conclusion, the integration of AI-assisted diagnosis with improved cytology screening techniques holds great promise for enhancing Cervical cancer detection and prevention. The proposed method, with its ability to learn from limited data and produce accurate diagnosis, offers a valuable contribution to the field of Cervical cancer screening.

\noindent\textbf{\large{Methods}}

\noindent\textbf{Efficient positive cell detection} \\
The role of the cell object detector is significant in two distinct aspects. 
Firstly, different from histological WSI where cells are small and densely packed\cite{campanella2019clinical,lu2021data}, the cells in cytology\cite{zeune2020deep,matek2019human} WSIs are comparatively larger. Thus a straightforward patching method used in histology may result in dividing a single cell into two separated patches, leading to the loss of crucial features. Moreover, lesion cells are often surrounded by normal cells, by focusing on the bounding boxes of individual cells rather than arbitrary patch division, precise and interpretable results can be obtained. 
Secondly, considering the semantic sparsity of CC WSI, the computational costs can be alleviated by top-K selection implemented with object detection. This approach effectively reduces the computational burden while maintaining adequate performance (checking the simple WSI-MIL theory foundation in the next subsection). Specifically, we perform object detection on the lesion or positive-like cells according to the TBS\cite{nayar2015bethesda} diagnostic classification but merge all positive categories for binary classification. 

To facilitate cell object detection, we preprocess the WSIs by capturing patch images with the size of 512x512 at a 10x magnification level. 
Since this process is for roughly lesion cell ranking, we utilize a small-size architecture YOLO-v5m\cite{jocher2020yolov5} as the detector to balance the speed and accuracy.

\noindent\textbf{SWIFT for both cell and WSI}\label{sec_wsi_method} \\
We first elaborate conventional approach for WSI analysis: Given a WSI $X$, the goal is to make slide-level prediction $\hat Y$ by learning a classifier $f(X;\theta)$.
Due to the extremely high resolution, in previous works\cite{campanella2019clinical, lu2021data, shao2021transmil, Zhang2022DTFDMILDF} $X$ is patched into a huge bag of small instances $X=\{x_1,..., x_N\}$, where N is the number of instances.
The slide-level supervision $Y$ is generally given by a Max-pooling operation of the latent label $y_i$ of each instance $x_i$, which can be defined as:
  \begin{equation}
    Y=\max\{y_1,..., y_N\},
    \label{EQ_wsi_maxpool}
  \end{equation}
where the class numbers should correspond to the degrees of lesion for multi-class subtyping. If all latent labels of instances $y_i$ are unknown (given only slide-level supervision), conventional approaches convert this problem into a MIL formulation in the following two steps:
1) Processing images into feature representations $Z=\{z_1,...,z_N\}$ with a backbone $h$ as $z_i=h(x_i;\theta_1)$ where $h$ is a CNN\cite{krizhevsky2012imagenet,he2016deep} or ViT\cite{dosovitskiy2020image,liu2021swin} model with pre-trained parameters $\theta_1$.
2) Aggregating all patches' features within a slide and producing the slide-level prediction $ \hat Y = g(Z; \theta_2)$,
where $g$ is an attention-based pooling function followed by a linear head for classification:
  \begin{equation}
      g(Z; \theta_2)= \sigma (\sum_{i=1}^N a_i z_i),
      \label{EQ_wsi_mil}
  \end{equation}
where $a_i$ is attention weights and $\sigma(\cdot)$ is a linear head.
Limited by the computational cost,
the parameters $\theta_1$ and $\theta_2$ in $f(X;\theta) = g\{h(X; \theta_1); \theta_2\}$ are learned separately by following steps: 
firstly initialize $\theta_1$ from the ImageNet pre-training\cite{lu2021data,NEURIPS2021_10c272d0}, or Self-SL\cite{li2021dual,chen2022scaling}, then freeze $\theta_1$ and learn $\theta_2$ under slide-level supervision.
A more common setting in real-world medical AI development is that we can obtain latent labels $\hat{y_i}$ of partial instances, thus representation backbone $z_i=h(x_i;\theta_1)$ with a cell-level classifier head $y_i=f(z_i;\theta_3)$ can be firstly learned via supervised training. Then, several methods can make WSI level diagnosis: directly predict by Max-pooling function\cite{campanella2019clinical} in Equation \eqref{EQ_wsi_maxpool}, or use WSI labels to further learn a MIL classifier by Equation \eqref{EQ_wsi_mil} based on fine-tuned $\theta_1$. The latter one method has been used in previous works on CC WSI Analysis system\cite{cheng2021robust,zhu2021hybrid} since cell-level classifier head may be biased during training with imbalanced data, and WSI architecture can add more complex non-linear operations for problem-solving.

Different to previous approach, our SWIFT method is trained end-to-end by utilize both cell- and WSI-level annotation informtion. The overall framework of SWIFT is illustrated in \textbf{Fig. \ref{fig_framework}b} and \textbf{Algorithm.\ref{alg_cell_wsi}}, which can be separated into the 3 components corresponding to the shown loss functions calculation pipeline:\\
1) The Semi-supervised Learning of cell classification:
given labelled cell images as supervised input stream and top-K cell images of a WSI as unsupervised input stream, they are trained via the way widely adopted Mean-Teacher\cite{tarvainen2017mean} paradigm of Semi-SL. 
The supervised stream are trained via strong augmentation on student model in a traditional way. 
In the meantime, the unsupervised stream first generates pseudo label via input weak augmentation into EMA teacher model, then input strong augmentation into student model to make prediction, thus the loss are calculated between the pseudo label and prediction.
In later works\cite{berthelot2019remixmatch,sohn2020fixmatch}, despite more complex input augmentations, the mask of pseudo label are proposed to simply refine the label via confidence threshold, e.g. filter out the pseudo label when the confidence is lower then 0.9.
\\
2) The weakly-supervised learning of WSI classification:
to address the issue of conventional method (the gap between pre-training and down-stream task), we propose to fine-tune the feature backbone and WSI head in an end-to-end mode based on the following hypothesis: the diagnosis of CC WSI via max-pooling can be transferred to relying on the minimal sufficient statistics of WSI, which is intuitive since pathologist usually find some supportive lesion cells to make final diagnosis. Denoting $P_{set} = \{p(\hat y_{1} | x_{1}), p(\hat y_{2}| x_{2}),..., p(\hat y_N | x_N)\}$ as a set of all instance's prediction within a WSI, then during inference Equation \eqref{EQ_wsi_maxpool} can be derived as:
\begin{equation}
    \hat{Y} = \max \{P_{set} \} = \max \{P_{subset} \} ,
    \label{EQ7}
  \end{equation}
where $P_{subset} \in P_{set}$, and represents top-K elements in $P_{set}$. The top-K operation can be realized by various visual attribution methods like the attention score of Transformer attention\cite{NIPS2017_3f5ee243}, CAM\cite{CAM, Grad-CAM}, LRP\cite{bach2015pixel}, Shapely\cite{chen2023algorithms} or Information Bottleneck\cite{tishby2000information,Li_2023_CVPR}. The successful condition of Equation \eqref{EQ7} is that the top-K recall is not zero. In this study, we get the top-K instances by sorting the probability of positive cells of object detector for its simplicity and better cell boundary. In addition, we can further distill top-K with a better cell classifier for higher top-K recall. 
\\
3) Incorporating above WSI-level weakly-supervised training into cell-level Semi-supervised Learning:
as illustrated above, the top-K cells of a WSI serve as the unsupervised stream of cell-level Semi-supervised Learning, thus they can be trained together and the gradient of both parts can be backward on cell encoder. 
But just to make cell pseudo labels more accurate, we further refine them via WSI-level information by the WSI-MIL head backward Grad-CAM of specific class.
Given top-K cells pseudo label and feature $\{ \hat{y}_k^{pseudo}, z_k \}_{k=1}^K $, the corresponding WSI-level label and prediction logits can be denoted as $Y$ and $\hat{Y} = g( \{z_k \}_{k=1}^K; \theta_2 )$. For the ground-truth class $c$ of $Y$, its activation map by Grad-CAM can be derived as:
\begin{equation}
 \beta_k^c = \frac{1}{ D }\sum_{d}^{D} \left( \frac{ \partial \hat{Y}^c}{\partial z_k^d} \right), 
 \label{eq_grad_cam}
\end{equation}
where $D$ is the dimension size of each feature instance $z_k$, and 
$\beta_k^c$ is the corresponding activation of WSI.
During labeling by pathologists, there is at least one specific category cell of a WSI, thus with WSI label information, the Grad-CAM based activation $\beta_k^c$ can be seen as cell-level pseudo label which contains WSI label information. 
So we simply threshold the CAM $\beta_k^c$ of all K instances, and repick them to refine cell-level semi-supervised pseudo label. 
Since CAM activation carries more information from WSI label, we set it with higher priority.

\begin{algorithm}
\caption{SWIFT}\label{alg_cell_wsi}
\renewcommand{\algorithmicrequire}{\textbf{Input:}}
\renewcommand{\algorithmicensure}{\textbf{Output:}}
\begin{algorithmic}[1]
\Require {a set of labelled cell images $D_{l}=\{ x_i,y_i\}_{i=1}^{N_{l}}$ and weakly labelled WSI bag of images $D_{w}=\{ X_j = \{ x_{jk} \} _{k=1}^{K}, Y_j \}_{j=1}^{N_{w}}$, where $N_{l}$ and $N_{w}$ represent the quantity of samples, and $K$ is the bag size of a WSI after top-K selection.}
\Ensure {a well-fine-tuned cell classifier and WSI classifier.}
\State {randomly initialize cell-level student and teacher model $h(x; \theta_1^s), h(x; \theta_1^t)$ and WSI-level model $f(X;\theta) = g(h(X; \theta_1^s); \theta_2)$}
\State {define $L_{cls}(\hat y_{notion},y_{notion}; m)$ as multi-classification loss with mask (default as no masking)}
\State {set batch size of cell and WSI dataloader as $b_c$ and $b_W$}
\For {each iteration} 
\State {randomly select $ \{ x_{i}, y_{i} \} _{i=1}^{b_c}, \{ X_{j}=\{ x_{jk} \} _{k=1}^{K}, Y_{j} \}_{j=1}^{b_W} $ data from $D_{l}, D_{w}$ to form training batch}
\State {strongly augment for all cell-level image: $x_{i}^{s}$, $x_{jk}^{s}$}
\State {weakly augment for weak supervision stream cell-level image: $x_{jk}^{w}$}
\State {cell prediction with student: $ \{\hat{y}_{i}, z_i ; \hat{y}_{jk} , z_{jk} \} = h( \{x_{i}^{s}; x_{jk}^{s} \}; \theta_1^s) $}

\State {WSI prediction: $ \hat{Y}_j = g ( \{ z_{jk} \}_{k=1}^{K} ; \theta_2 ) $}
\State {pseudo label generation with teacher: $ y_{jk}^{pseudo} = f(x_{jk}^{w};\theta_1^{t}) $}
\State {obtain CAM activation $\beta _k^c$ by Equation. \eqref{eq_grad_cam}.}
\For {k=1:K}
\If{$\beta _k^c > \tau_1 $}
        \State {$m_{jk} = 1$, $y_{jk}^{refine}=c$, \# set the corresponding category}
\ElsIf{$ \max \{ y_{jk}^{pseudo} \} > \tau_2 $, \# find max probability } 
\State {$m_{jk} = 1$, $y_{jk}^{refine}= \arg \max \{ y_{jk}^{pseudo} \}$}
\Else {}
\State {$m_{jk} = 0$, \# omit $y_{jk}^{refine}$ here since it will be masked in loss}
\EndIf
\EndFor
\State {semi-weakly supervised loss: $L_{semi\_weak}= L_{cls}(\hat{y}_{jk},y_{jk}^{refine}; m_{jk})$ \# mask ref}
\State {cell supervised loss: $L_{cell\_sup}= L_{cls}(\hat{y}_{i},y_i)$}
\State {WSI supervised loss: $L_{wsi\_sup}= L_{cls}(\hat{Y}_{j},Y_j)$}
\State {overall loss: $L_{total}=L_{cell\_sup}+ L_{wsi\_sup} + L_{semi\_weak} $}
\State {backward loss gradient and update student $\theta_1^s$ and WSI head $\theta_{2}$}
\State {update teacher params $\theta_1^{t}$ by exponential moving average of $\theta_1^{s}$}
\EndFor

\end{algorithmic}
\end{algorithm}

\noindent\textbf{Explainable cell classification assisted by diagnostic description prediction}\\
Previous works primarily focus on generating descriptions similar to image captioning, which can be difficult due to the complexity and diversity of both images and their descriptions, especially with limited pathology data. To tackle this issue, we propose utilizing a set of $N$ predefined cell diagnostic description options to constrain the training process, as illustrated in \textbf{Extended Data Table. 2}. Essentially, this module is responsible for selecting appropriate descriptions from the predefined set, which contains the following three parts to ensure a more reliable description prediction. The overall structure of this module is shown in \textbf{Fig. \ref{fig_framework}d}.
\\
1) Cell \&Description Alignment. To begin, we employ a text encoder (Roberta-base\cite{liu2019roberta}) to extract predefined description embeddings $T$, and an image encoder that is trained in this study to extract cell feature $V$, then we align the cell feature  $V$ with its corresponding descriptions by learning a multimodal embedding space, similar to the approach used in CLIP\cite{radford2021learning}. Specifically, we maximize the cosine similarity between the cell feature and its corresponding $n$ descriptions among a set of $N$ predefined descriptions, while minimizing the cosine similarity with the remaining $N-n$ unrelated descriptions. Since the image-text relationship is bidirectional, we optimize the model by minimizing a symmetric binary cross-entropy loss $L_{bce}$, which is formulated as follows:
\begin{equation}
	\label{eq:predict}
	\ell _{align}=L _{bce}(\sigma (\frac{VT^{\top }}{\parallel {V\parallel\parallel T \parallel}} ), Y_{d}) +L _{bce}(\sigma (\frac{TV^{\top }}{\parallel {V\parallel\parallel T \parallel}} ), Y_{d}^{\top}),
\end{equation}
where $\sigma$ denotes the Sigmoid function, $Y_{d}  \in  \mathbb{R} ^{bs*N}$ denotes the multi-hot text description label corresponding to the cell in a training batch $bs$. It is also worth noting that, in order to maintain the cell classification performance, we freeze the image encoder and only finetune the text encoder.
\\
2) Valid Description Selection. The part is designed to select valid descriptions $T_{p}$ from predefined $N$ descriptions, where a description is deemed valid if its cosine similarity $sim_{i}$ to a given cell representation exceeds threshold $\lambda$, formulated as:
\begin{equation}
	\label{eq:predict2}
	T_{p} = \{{T_{i} \mid }  sim_{i}> \lambda, i\in [1,N] \}, where \ sim_{i}=\frac{VT_{i}^{\top }}{\parallel {V\parallel\parallel T_{i} \parallel}}
\end{equation}
And we use the predicted descriptions $T_{p}$ to represent the explanation of cell classification.

\begin{algorithm}
\caption{Diagnostic Description Prediction}\label{alg_exp_cell2}
\begin{algorithmic}[1]
\Require
A set of labelled images with their descriptions \(D_{l}=\{ x_{i},y_{i}\}_{i=1}^{N_{l}}\), where \(N_{l}\) is the number of samples, \(x\) denotes the input image and \(y\) denotes the corresponding description.
A trained image encoder \(f(X;\theta_{img})\) with frozen parameters.
\Ensure A well-fine-tuned description selection model and  predicted descriptions.

\State Initialize a pre-trained text encoder (e.g., Roberta-base) as \(f(X;\theta_{text})\).

\Statex{\textbf{\# Training Phase}}
\For {each iteration} 
    \State Randomly select a batch of \( \{ x, y \} \) from \(D_{l}\).
    \State Extract cell features \(V\) using the image encoder: \(V = f(x;\theta_{img})\).
    \State Extract description features \(T\) using the text encoder in inference mode: \(T = f(y;\theta_{text})\).
    \State Align the cell feature \(V\) with its corresponding descriptions using symmetric binary cross-entropy loss \(L_{bce}\):
    \[
    \ell_{align}=L_{bce}\left(\sigma\left(\frac{VT^{\top}}{\|V\|\|T\|}\right), Y_{d}\right) + L_{bce}\left(\sigma\left(\frac{TV^{\top}}{\|V\|\|T\|}\right), Y_{d}^{\top}\right)
    \]
    \State Update the text encoder parameters \(\theta_{text}\) by back-propagating the alignment loss \(\ell_{align}\).
\EndFor

\Statex{\textbf{\# Inference Phase}}

\State Compute the cosine similarity \(sim_{i}\) for all descriptions:
\[
sim_{i}=\frac{VT_{i}^{\top }}{\|V\|\|T_{i}\|}
\]
\State Select valid descriptions \(T_{p}\) based on the threshold \(\lambda\):
\[
T_{p} = \{T_{i} \mid sim_{i}> \lambda, i\in [1,N] \}
\]
\State Use the predicted descriptions \(T_{p}\) to represent the explanation of cell classification.
\end{algorithmic}
\end{algorithm}



\clearpage

\noindent\textbf{Color adversarial training} \\
The color adversarial training (ColorAdv) is proposed to enhance the model's robustness to the color variants. Different to conventional adversarial robustness\cite{rocher2023adversarial,woods2019adversarial}, we mainly consider the color space of input for pathology data reality. For simplicity, considering only the slide-level end-to-end training, the ColorAdv can be formulated as a min-max process:
	\begin{equation}\label{eq1}
		\min_{\theta} \max_{ r, || r||<\rho}; L_{cls}( X+ r,  Y; \theta) + L_{cls}( X,  Y; \theta), 
	\end{equation} 
where $r$ is a three-dimensional variable and is used to modify the color values of the WSI slide, but bounded by a threshold hyper-parameter $\rho$ to control the amplitude of the color adversarial perturbation, and $\theta$ corresponds to $\{\theta_1, \theta_2\}$ for cell backbone and WSI head. The ColorAdv can be divided into two steps,  color adversarial example generation ($ r_{adv} := \arg\max_{ r};|| r||<\rho L_{cls}( X+ r,  Y; \theta)$) and generalized model learning ($\min_{ \theta} L_{cls}( X+ r,  Y; \theta) + L_{cls}( X,  Y; \theta)  $). These two steps are performed in one training iteration by an alternative training strategy.  The overall framework is illustrated in \textbf{Fig. \ref{fig_framework}c.}
 \\
1) Color adversarial example generation. In this step, the color adversarial perturbation $ r_{adv}$ is calculated by maximizing the network’s prediction error. We can approximate $ r_{adv}$ with a linear approximation of loss with respect to $ r$. More precisely, it can be formulated as
	\begin{equation}\label{eq2}
		 r_{adv} \approx \rho \frac{ g}{|| g||_2}, \quad  
		\text{given} \quad  g = \nabla_{ r} L_{cls}( X+ r,  Y; \theta)
	\end{equation}
After calculating the $ r_{adv}$, the color adversarial examples are generated by $ X +  r_{adv}$.
 \\
2) Generalized model learning. After generating color adversarial examples, the model parameters are updated by training on a mixture of generated color adversarial and original examples.
\\
3) Random color space choice.
Diversity and realism of data augmentation are critical for mitigating the domain shift between augmented and test data and further improving the model's generalization ability\cite{xie2020unsupervised}, which is explored from the perspective of color space choice in this section.  Injecting the color adversarial perturbation on different color spaces can help to diversify the style of the augmented data since they modify the color in a different way. Accordingly,  performing ColorAdv on the combined color space is adopted to further promote model performance. Specifically, two hyper-parameters, $\rho_{rgb}=0.1$ and $\rho_{hsv}=0.1$, are set to control the amplitude of the color adversarial perturbation for the RGB and HSV spaces, respectively.

\noindent \textbf{Dataset}

The dataset is meticulously annotated and curated by pathologists through our custom web-based annotation programs, following rigorously designed procedures. For a detailed overview of the data preparation process and a concise summary of dataset information, please refer to \textbf{Fig. \ref{fig_datasets1}} and \textbf{Extended Data Fig. 1}.
We collect our Cervical cytology Liquid-based Prepared smear slides dataset from multi medical centers or hospitals. These slides are scanned with a ×40 magnification (0.25 $\mu m$ per pixel resolution). 
Based on the diagnosis report from the collected hospital of each WSI, 3 professional pathologists further reviewed the the WSI category according to the TBS diagnostic criteria. The digital slides are classified into 6 different categories (\textbf{Extended Data Fig. 1c}) by manual annotation of pathologists using our web-based annotation software (\textbf{Extended Data Fig. 1a}), which provides visualizing, annotating function of WSIs.
After removing unqualified slides via Qualify Control like image clarity and cell number requirements, a total of 106,426 WSIs are obtained for model development and the average slide dimension (height×width) was 74890.1998±26336.4773 × 75455.2696±24637.4448 (mean±standard deviation).
Among these, 22,601 WSIs (21.2\%) are annotated as lesion-positive, while the remaining 83,825 WSIs (78.8\%) are negative. Among the subtype of lesion-positive WSIs, about 71.6 \% WSIs are low-grade lesion (47.8 \% of ASC-US and 23.8 \% of LSIL), and remaining are classified as high-level lesions or gland lesion. Notably, each WSI represents a unique patient case, with a diverse and representative collection of cell instances for analysis.
The large volume cells and high resolution with in a WSI not only makes comprehensive examination time-consuming for pathologists, but also leads difficulty for effective model training.
Unlike histopathology WSIs where lesion cells tend to form distinct tumor regions, the isolated arrangement of cells in cytology WSIs makes it considerably more demanding to achieve precise and exhaustive annotations. What's more, the scarcity of domain-expert pathologists further compounds the challenge of expeditious and accurate labeling.

To annotate CC WSIs at the cell level, pathologists must meticulously scan small patches of each WSI to identify lesion cells. However, relying solely on the expertise of pathologists becomes insufficient to meet the growing demand for annotating the expanding dataset. For instance, during our development, one pathologist could annotate only about 20 WSIs per day, while we had over 100,000 WSIs. As a result, pathologists were primarily assigned to label the initially collected 10,000 WSIs at both the cell and WSI levels.
Using the annotated cells, a cell model is trained to assist in a simple active learning paradigm\cite{radsch2023labelling,prince2004does,budd2021survey}. Pathologists modify false-positive cells among the top-K predictions, after which the remaining WSI data are primarily annotated at the WSI level and adjusted at the cell level. This stratified approach is necessary due to the increasing volume of collected data and the limitations of pathologist resources. Despite the stochastic nature of the annotation process, the collected cell annotation quantity remains limited (roughly 200k positive cell annotations / (22,623 positive WSIs * top-256) $\approx$ 3.44\%). In other words, less than 3.5\% of positive cells are annotated. Consequently, innovative techniques are needed to alleviate the annotation demand for training a high-performance AI diagnosis model.
For textual description data of cells, pathologists are asked to select the corresponding pre-defined description or manually input during above cell annotation process.

\noindent \textbf{Data split for training and evaluation.}
For the 65 centers of WSIs data, we firstly random select about 8k data for testing and  randomly split the remain data into 8:2 for training:validation. Specially, to avoid data leakage of cells from WSIs, we only use cells or patches sampled from training WSI to construct train-val data in cell detection and cell classification training, and use the cells of remaining testing WSI for cell detector or classifier testing.  
To assess the robustness of our proposed method to domain-shift, we designated 10 centers with unseen domain data as an additional testing set. This is crucial for real-world AI deployment, an aspect where previous works\cite{cheng2021robust,zhu2021hybrid} ignored but struggled to generalize. The details of this data can be refereed to \textbf{Fig. \ref{fig_datasets1} and \textbf{Extended Data Fig. 1}}.
For real-world external test data, the AI model are deployed to 105 centers with human pathologist's diagnosis review. Totally 234,846 WSI (10614 positive patient) are tested (see \textbf{Fig. \ref{fig_datasets1}b}, \textbf{Fig. \ref{fig_5_external}a+b} and \textbf{Extended Data Fig. 1} for more details).
For the clinical trials on 3 hospitals (1954 WSI with 212 positive), the trial's workflow is listed in \textbf{Extended Data Fig. 2}, mainly including participated patient and slide imaging control, controlled trial on if assisted by our AI model, third-party expert for gold standard labelling and data statistical analysis. 
Additionally, One of trials record are partially listed in \textbf{Extended Data Table. 1}. For each centers' data details, the confusion matrix in \textbf{Extended Data Fig. 3} can be refereed.

\noindent \textbf{Performance evaluation.}
Noting that most pathologists / doctors care more about the WSI diagnosis result compared to cells performance, we claim that the WSI performance is more important also considering that the cells can only be sparsely annotated. Since doctors also mainly care about sensitivity and specificity of binary classification, thus for WSI, we choose AUC (Area Under ROC Curve) and specificity at sensitivity=95 (by tuning softmax probability threshold) as the main measurement metric during model development. Our AUC mainly focus on binary result, thus we sum all positive categories' probability for binary prediction logit. For model evaluation on external test data and clinical trial, the direct sensitivity and specificity are reported. To explore the fine-grained classification on imbalance WSI dataset, we also add average F1-macro score for additional evaluation. The performance of different methods was compared by the average of above metrics of three iterations.
For lesion-positive cells detection, we choose COCO mAP@50-95 (mean Average Precision ranging from IoU=50 to 95, stride=5) as metric (some high level lesion cells are too small thus too high IoU may miss-lead us). Since this detector only works as a bounding-box regressor and a rough positive cell ranker, we do not make too much evaluation and thorough parameters tuning.
For fine-grained multi-class cell classification, the average F1-macro score is used for evaluation since the category distribution is strongly imbalanced.

\noindent \textbf{Compared baselines.}
We compare our method with the following baselines including unlabelled Self-supervised pre-training: SimCLR\cite{chen2020simple}, MoCo\cite{MOCO}, DINO\cite{caron2021emerging}, as well as Semi-supervised Learning methods 
including Mean-Teacher\cite{tarvainen2017mean}, Fix-Match\cite{sohn2020fixmatch}, Flex-Match\cite{zhang2021flexmatch} and Semi-supervised Learning for imbalance data method Crest\cite{wei2021crest}. 
To make comparison of WSI level architecture, we also include some methods proposed in histological WSI classification: ABMIL\cite{ilse2018attention}, TransMIL\cite{NEURIPS2021_10c272d0}. 
We also extract features and use simple Mean / Max-pooling to form MIL classifier training.
Similar to the linear probing method in Self-SL\cite{chen2020simple, MOCO}, we also compared the frozen  pre-trained features (extract by Semi-/Self-SL and full supervised training on cell level data) and fine-tuned features for WSI diagnosis.

For backbone selection, we mainly compared ConvNext-tiny and ViT-small, and use ViT-small for Self-SL in DINO\cite{caron2021emerging}. Since Batch Normalization shows training eval gap in ResNet, we find that it is not work well (collapse) when fine-tuning with WSI top-K where all patches generally share similar style or statistics within a WSI. The Layer Normalization used in ConvNext and ViT show advantages in such sequence modeling, similar to the NLP tasks. 

\noindent \textbf{Implementation details.}
We implement PyTorch library and 4 A100-40g GPUs for parallel training via PyTorch DistributedDataParallel.
We use the training set to optimize all above models, the validation set to adjust the hyper-parameters, and then test the performance on the testing set. The 10 domains unseen dataset is used for evaluating model generalization or robustness to staining or imaging changes. The positive and negative cell samples are cropped around the annotations of positive slides, then the negative samples are randomly cropped from negative slides for further supplement.
For object detection, we follow the most settings in YOLO-v5\cite{jocher2020yolov5}, the backbone model is chosen as small size. Since strong augmentations can easily hurt cells' detailed key features, we choose augmentation parameters with low degree. The model is trained with a batch size of 64, where the learning rate is initialized as 1e-2 and gradually linear decays with 300 epochs.

The SWIFT model is optimized by Adam optimizer with an initial learning rate of 1e-3 and weight decay as 1e-3, the learning rate is scheduled with linear decay of about 320k iteration steps. 
We use a global Attention module with hidden feature size of 128, and add dense connected MLPs with 4 layers (each followed with a Layer Normalization layer) to improve WSI-level representation ability for classifying WSIs. To avoid over-fitting of WSI-Attention module, the Dropout is added with mask ratio of 0.1.
In this training process, data augmentation is added including rotation, horizontal and vertical flips, and random crop, also the random rotation and random HSV color-jitter. During training, the data loader of labelled cells with a batchsize of 128 and unlabelled top-256 cells of 2 WSIs (bag size = 256, WSI-level batchsize = 2) are concatenated as a tensor to input into the cell encoder parallel. To save memory usage, mix-precision training is employed by PyTorch Autocast. 
After convergence, the model is further fine-tuned via color adversarial training to improve domain change robustness.
For the diagnostic description prediction module, we add a fully connected layer on top of our trained patch-based image encoder to map features to diagnostic descriptions, thereby generating corresponding cell-level description predictions. We freeze the image encoder, employ a learning rate of 1e-4, and use the AdamW optimizer to train the model over 10 epochs.

\end{spacing}

\clearpage

\begin{figure*}  
\centering
\includegraphics[width=0.9\textwidth]{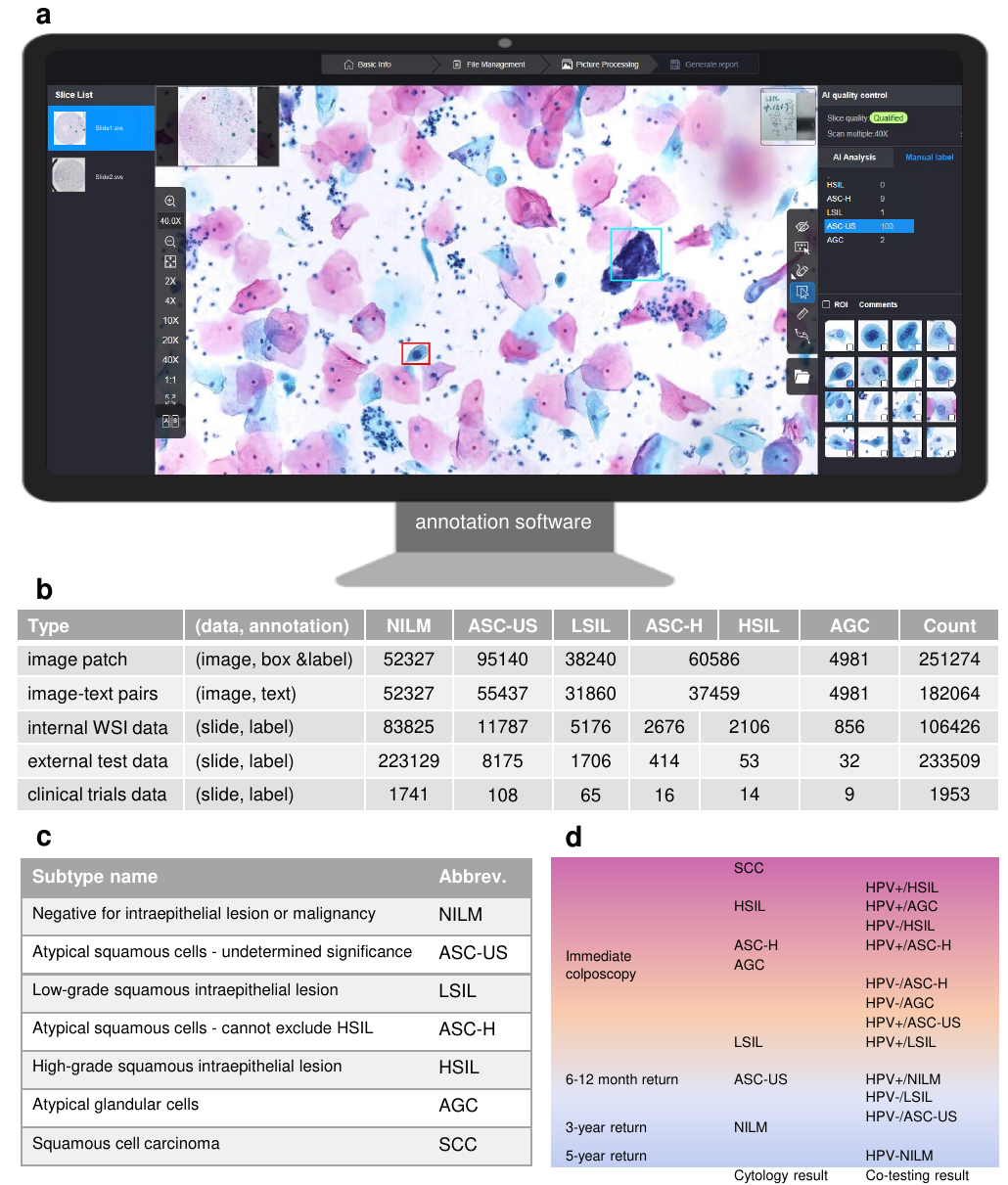} 
\caption*{\textbf{Extended Data Figure 1. Overview of our dataset construction. 
a.} Illustration of the data annotation using our developed web program. 
\textbf{b.} Summary of each organized data set.
\textbf{c.} The lesion subtype in TBS criteria and its abbreviation. \textbf{d.} The subtype distribution in feature space, which is in line with the model prediction confusion matrix.}  
\label{fig_datasets}
\end{figure*}

\begin{figure*}
    \centering
    \includegraphics[width=\textwidth]{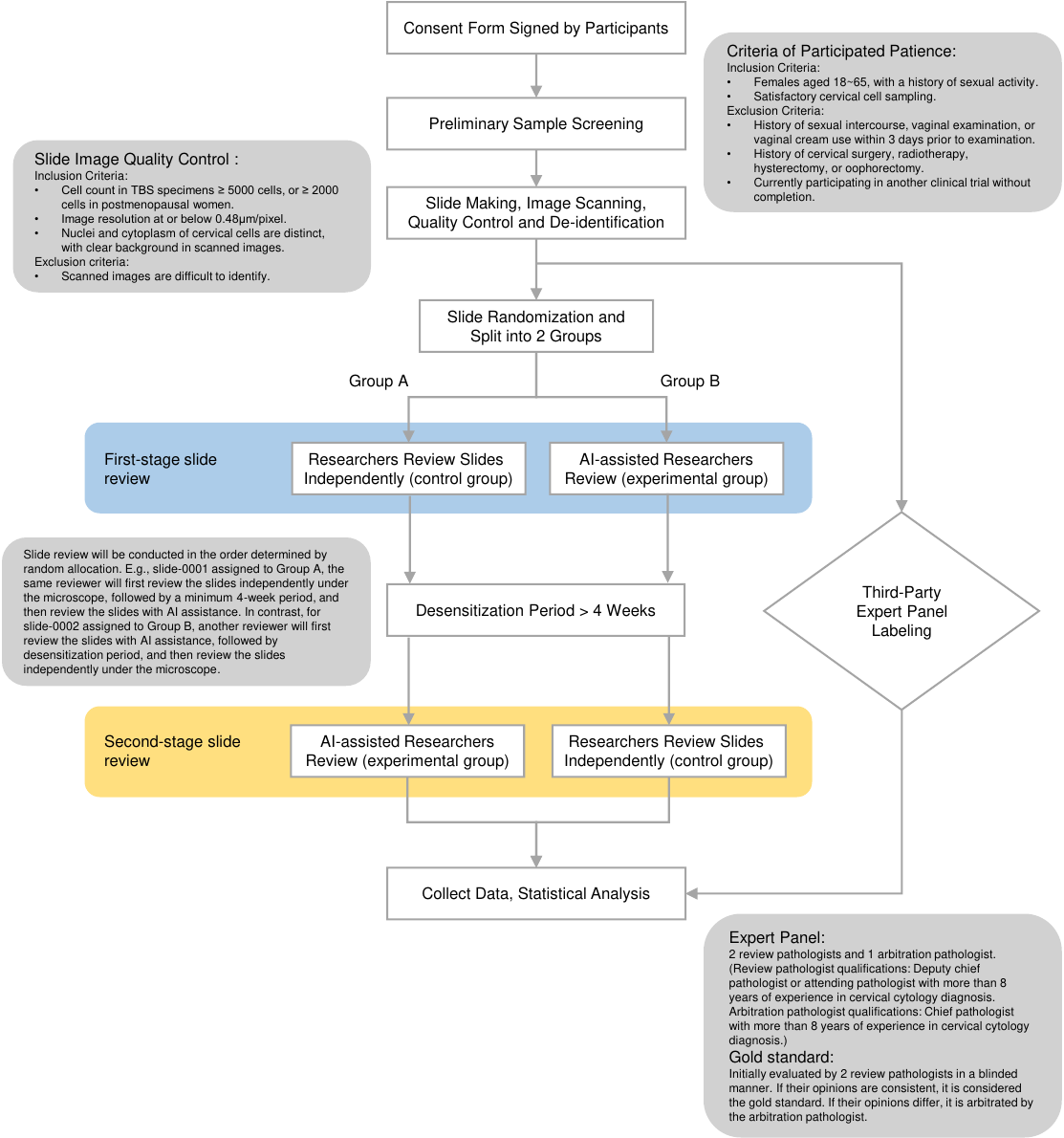}
    \caption*{\textbf{Extended Data Figure 2.} Overview of the workflow of the  real-world Clinical Trial study. }  
    \label{fig_clinical_trial_sop}
\end{figure*}

\begin{figure*}
    \centering
    \includegraphics[width=0.8\textwidth]{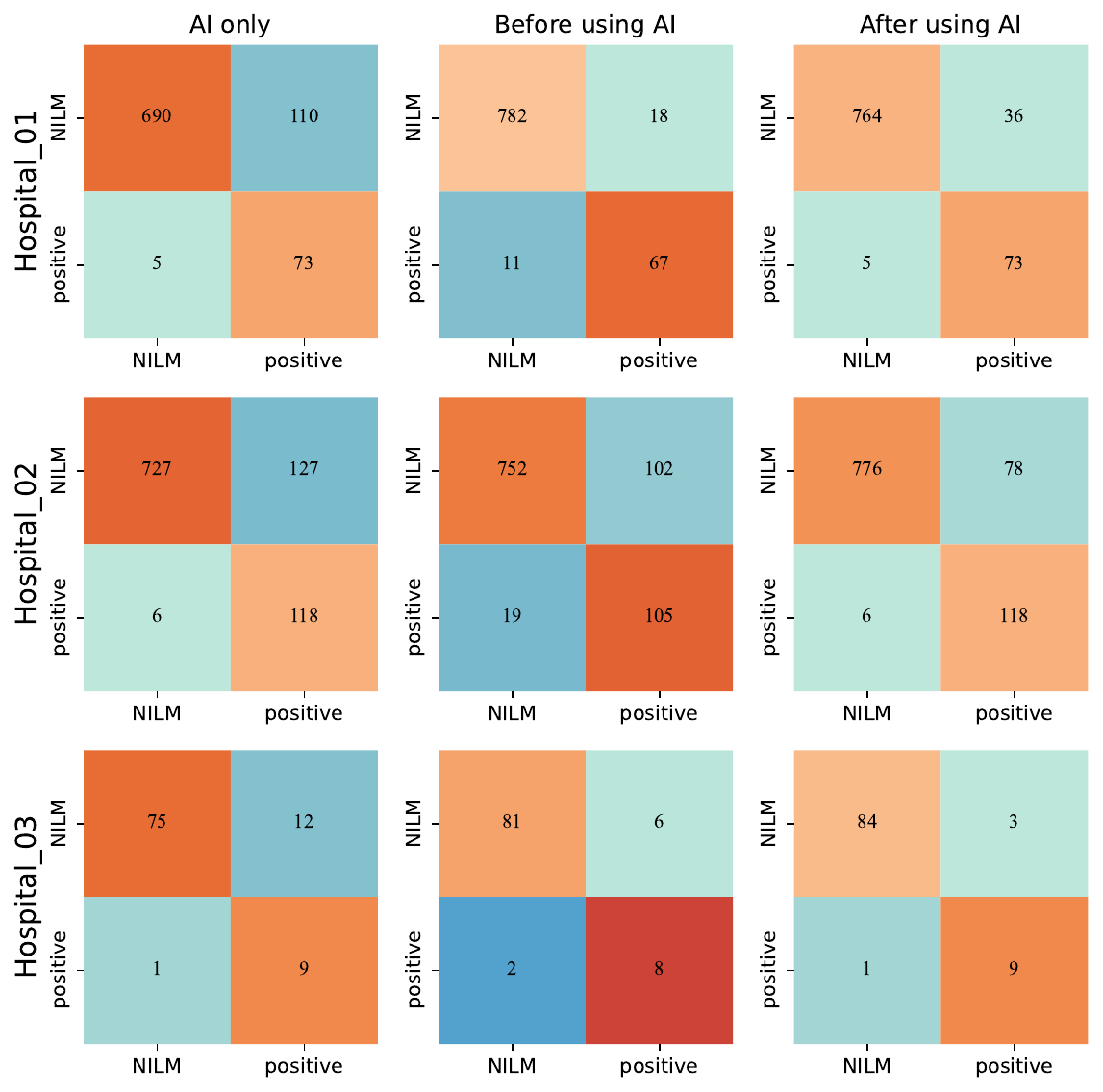}
    \caption*{\textbf{Extended Data Figure 3. Clinical Trial confusion matrix of three hospitals with strong ability and reputation. 
    }  }
    \label{fig_clinical_trial_sop}
\end{figure*}

\begin{figure*}
    \centering
    \includegraphics[width=0.8\textwidth]{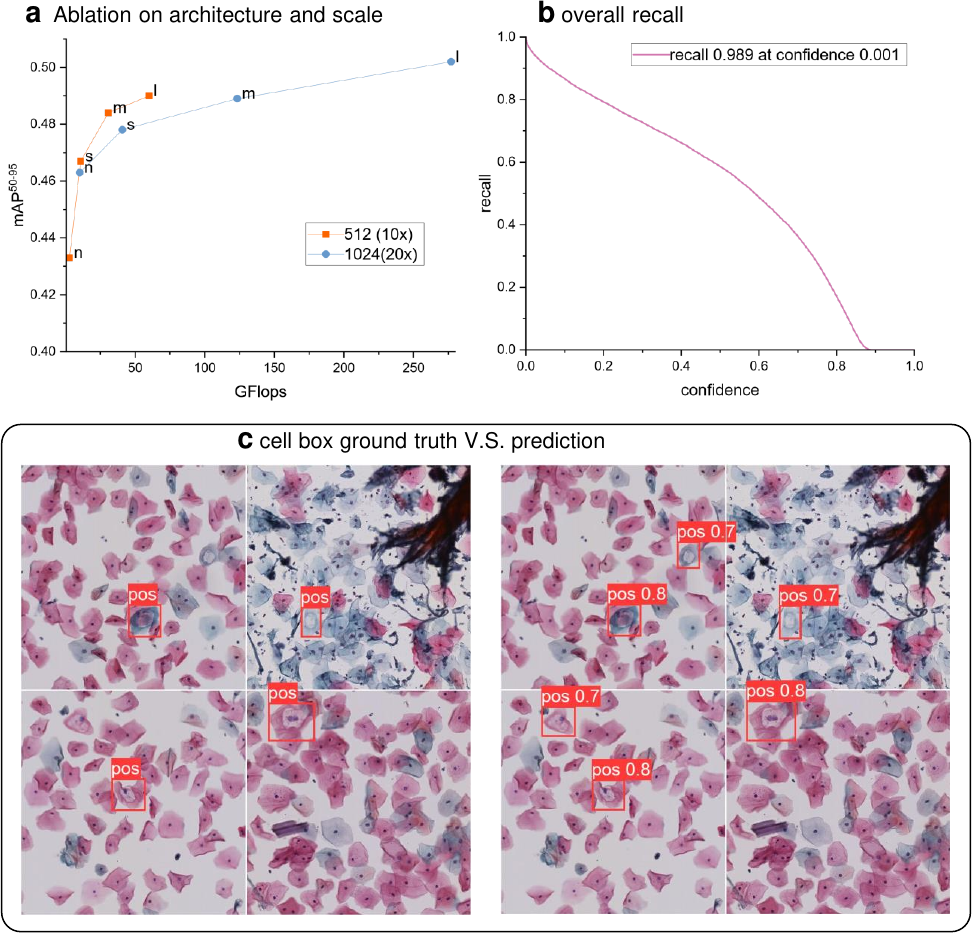}
    
    \caption*{  \textbf{Extended Data Figure 4.} \textbf{Positive cell detection.}
    \textbf{a.} The ablation on architecture and scale reveals that we can use YOLO-v5m on 10x magnification scale to detect cells for effectiveness and efficiency. \textbf{b.} The overall recall reaches 0.989 with a small confidence threshold of detector, which is hardly to miss positive cells within WSI. \textbf{c.} The detector can find almost all positive annotated cells and also with strong generalization to detect some overlooked positive-like cells.
    }  
    \label{fig_cell_od}
\end{figure*}

\begin{figure*}
    \centering
    \includegraphics[width=0.85\textwidth]{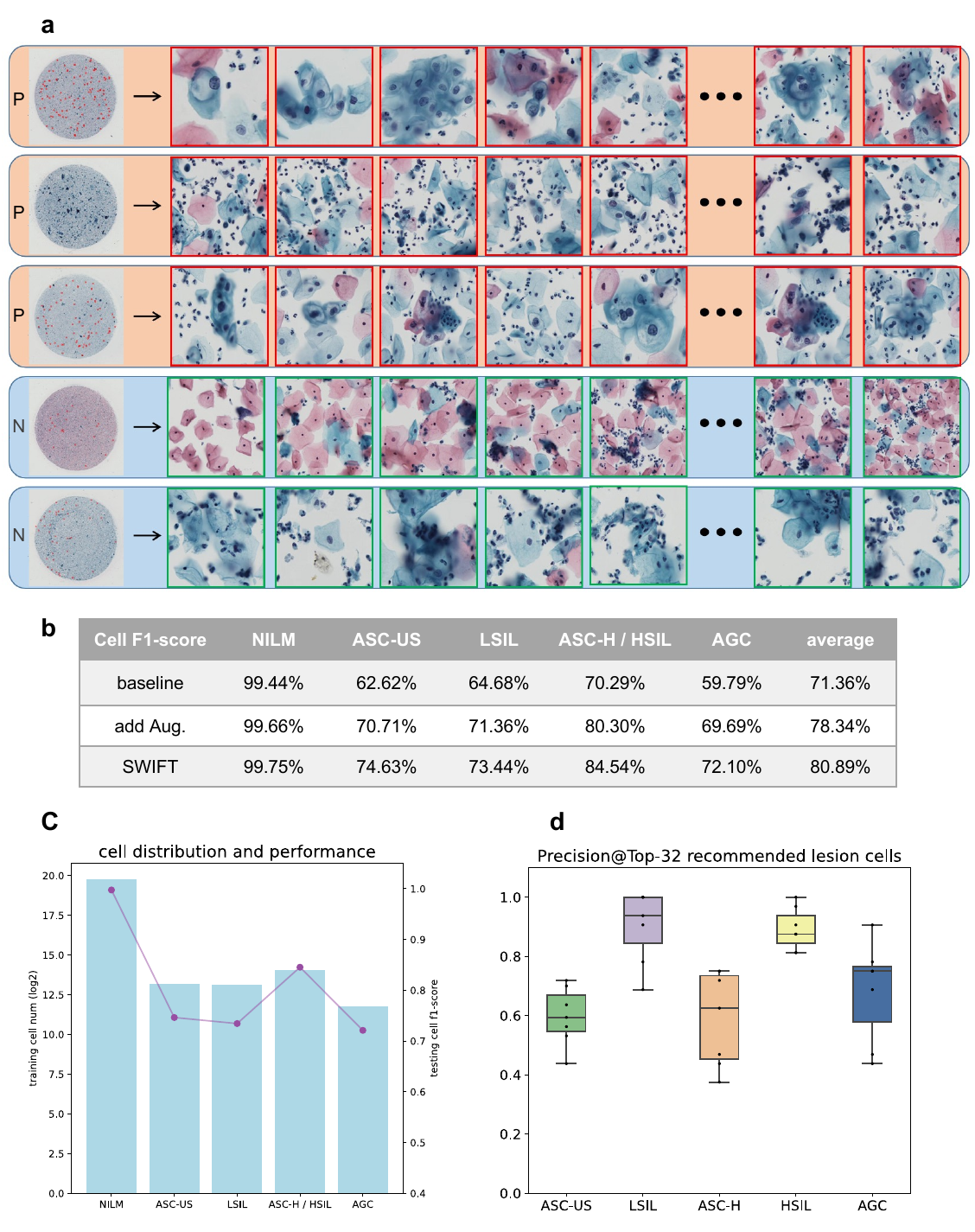}
    
    \caption*{  \textbf{Extended Data Figure 5.} \textbf{Cell classification.}
    \textbf{a.} The predicted top cells (or Region of Interest, ROI) of both positive and negative WSIs, based on these ROIs, pathologists can make diagnosis more quickly without scanning the whole image. \textbf{b.} Our SWIFT model combine WSI-level information can boost the cell classification compared to baseline or add augmentations (Aug.). More importantly, it can smooth the mismatch problem of WSI-label and cell-level prediction, e.g. WSI labelled as AGC but cell classifier predicted ROIs are all ASC-US. \textbf{c.} Cell-level performance are highly correlated to the quantity of each categories. \textbf{d.} Precision@Top-32 ROIs of WSI, note that for different WSI category, it's ROIs precision are varied. For LSIL and HSIL WSIs, it generally contains a lot of positive cells which could be easily distinguished. For other types (Atypical ...), the precision is likewise consistent with their confused feature.
    }  
    \label{fig_cell_cls}
\end{figure*}

\begin{figure*}
\centering
\includegraphics[width=\textwidth]{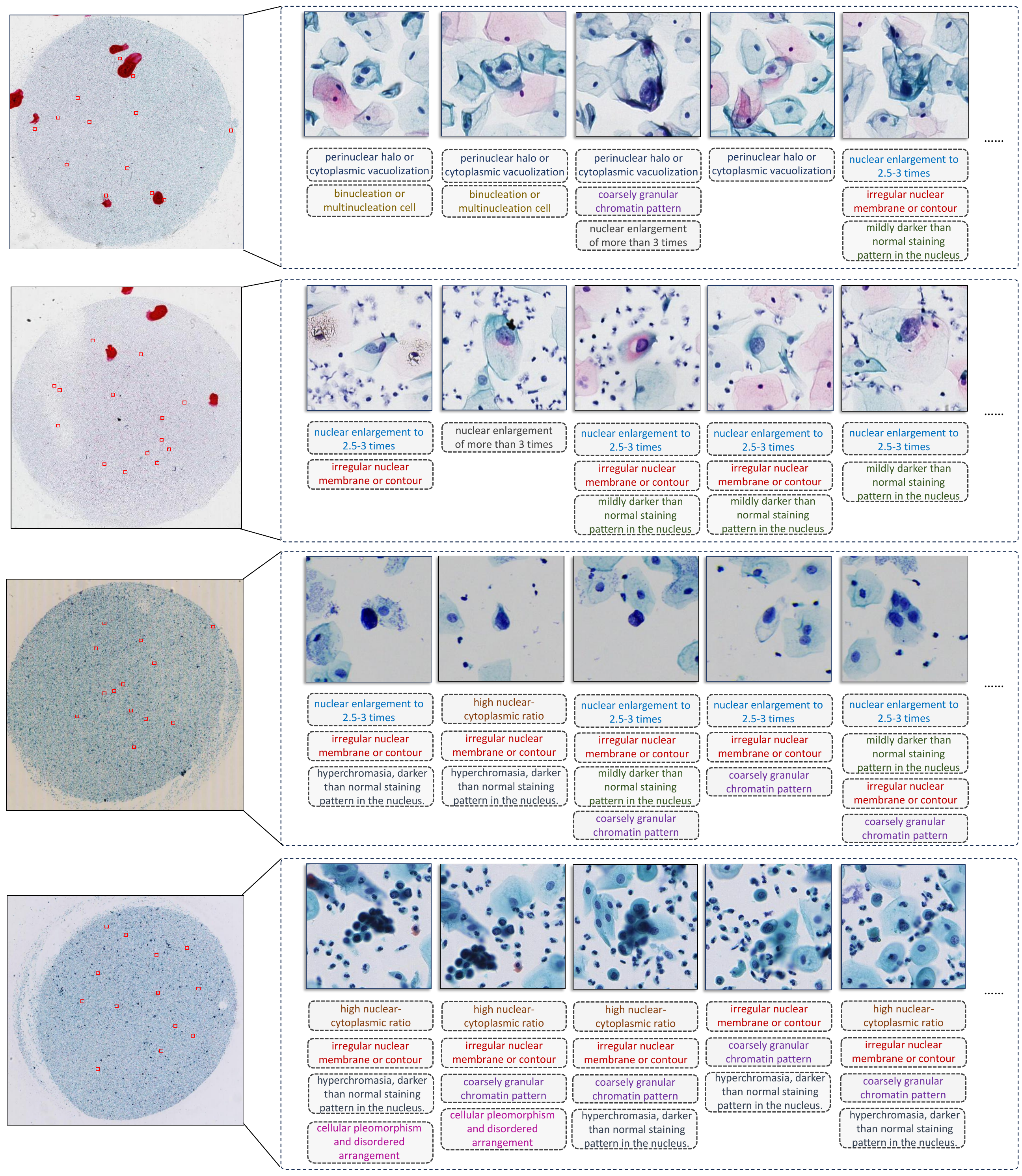}  
\caption*{  \textbf{Extended Data Figure 6.} Examples of interpretable classification of cells in WSI.}  
\label{fig_cells_method2}
\end{figure*}

\clearpage

\begin{table}
    \centering
    \includegraphics[height=1.2\textwidth]{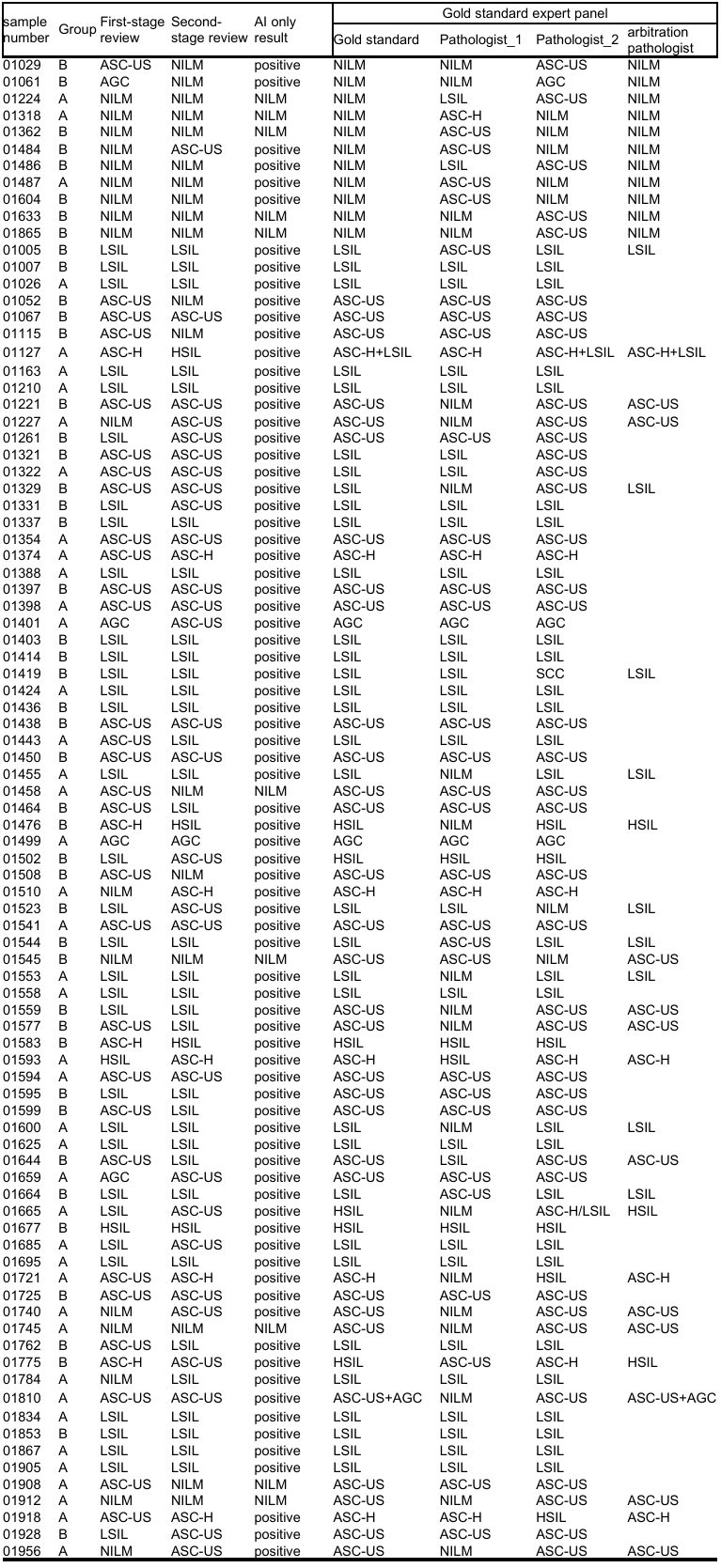}
    \caption*{\textbf{Extended Data Table 1. Clinical Trial Data Record}. Here we show all the positive samples and samples need arbitration, from Hospital\_01. \textbf{Key-point: a.} There is no lesion patient is missed by both AI and human reviewers when lesion conditions severe than ASC-US. \textbf{b.} After arbitration, some lesion-positive slide suspected by one human reviewers are determined as NILM (negative) diagnosis.}  
    \label{table_clinical_trail_data}
\end{table}

\begin{table}
    \centering
    \begin{tabular}{|c|l|c|}
        \hline
        \textbf{Cell type} & \textbf{Description} & \textbf{Number} \\
        \hline
        \multirow{1}{*}{NILM} & normal cell & \cellcolor{N!100}52373 \\
        \hline
        \multirow{5}{*}{ASC-US} & nuclear enlargement to 2.5-3 times & \cellcolor{ASCUS!44}23024 \\
        & irregular nuclear membrane or contour & \cellcolor{ASCUS!26}13545 \\
        & mild hyperchromasia, mildly darker than normal staining pattern in the nucleus, & \cellcolor{ASCUS!30}15699 \\
        & coarsely granular chromatin pattern & \cellcolor{ASCUS!6}3169 \\
        
        \hline
        \multirow{8}{*}{LSIL} & nuclear enlargement of more than 3 times & \cellcolor{LSIL!6}3138 \\
        & binucleation or multinucleation cell & \cellcolor{LSIL!7}3699 \\
        & coarsely granular chromatin pattern & \cellcolor{LSIL!6}3308 \\
        & irregular nuclear membrane or contour & \cellcolor{LSIL!13}6692 \\
        & marked nuclear dyskaryosis (raisinoid nuclei) & \cellcolor{LSIL!3}1281 \\
        & perinuclear halo or cytoplasmic vacuolization & \cellcolor{LSIL!26}13547 \\
        & hyperchromasia, darker than normal staining pattern in the nucleus. & \cellcolor{LSIL!1}195 \\
        
        \hline
        \multirow{6}{*}{ASC-H / HSIL} & high nuclear-cytoplasmic ratio & \cellcolor{ASCH!16}9070 \\
        & hyperchromasia, darker than normal staining pattern in the nucleus. & \cellcolor{ASCH!20}10726 \\
        & irregular nuclear membrane or contour & \cellcolor{ASCH!18}9762 \\
        & coarsely granular chromatin pattern & \cellcolor{ASCH!10}5903 \\
        & Cellular pleomorphism and disordered arrangement & \cellcolor{ASCH!3}1998 \\
        \hline
        \multirow{6}{*}{AGC} & prominent single or multiple nucleoli & \cellcolor{AGC!8}413 \\
        & hyperchromasia, darker than normal staining pattern in the nucleus. & \cellcolor{AGC!27}1418 \\
        & nuclear pleomorphism, variation in size and shape of nuclei & \cellcolor{AGC!25}1309 \\
        & irregular nuclear membrane or contour & \cellcolor{AGC!16}823 \\
        & cytoplasmic vacuolation & \cellcolor{AGC!1}40 \\
        & cell group with rosette or acinar formation with feathering & \cellcolor{AGC!3}129 \\
        & loss of nuclear polarity and disorganized arrangement & \cellcolor{AGC!16}811 \\
        & upward nuclear displacement & \cellcolor{AGC!1}38 \\
        \hline
    \end{tabular}
    \caption*{ \textbf{Extended Data Table 2. Visualization of various cell descriptions.} The first column is the cell type, the second outlines possible diagnostic interpretations, and the third is their corresponding number of annotations. }  
\label{table_cell_des}
\end{table}

\clearpage
\section*{References} 
\vspace{2mm}

\begin{spacing}{0.9}
\bibliographystyle{naturemag}
\bibliography{main}
\end{spacing}

\end{document}